# A Framework for Human Evaluation of Large Language Models in Healthcare Derived from Literature Review

**Thomas Yu Chow Tam, MS[1*], Sonish Sivarajkumar, BS[2*], Sumit Kapoor, MD[3], Alisa V Stolyar[1], Katelyn Polanska, BS[1], Karleigh R McCarthy, BS[1], Hunter Osterhoudt, MS[1], Xizhi Wu, MS[1], Shyam Visweswaran, MD, PhD[4], Sunyang Fu, PhD[5], Piyush Mathur, MD[6], Giovanni E. Cacciamani, MD[7], Cong Sun, PhD[8], Yifan Peng, PhD[8], Yanshan Wang,PhD[1,2,4,9,10 †]**

*[1]Department of Health Information Management, University of Pittsburgh, Pittsburgh, PA, USA; [2]Intelligent Systems Program, University of Pittsburgh, Pittsburgh, PA, USA; [3]Department of Critical Care Medicine, University of Pittsburgh Medical Center, Pittsburgh, PA, USA;[4]Department of Biomedical Informatics, University of Pittsburgh, Pittsburgh, PA, USA; [5]Department of Clinical and Health Informatics, Center for Translational AI Excellence and Applications in Medicine, University of Texas Health Houston, Houston, TX, USA; [6]Department of Anesthesiology, Cleveland Clinic, Cleveland, OH, USA; [7]Department of Urology, Keck School of Medicine, University of Southern California, Los Angeles, CA, USA; [8]Department of Population Health Sciences, Weill Cornell Medicine, New York, NY, USA; [9]Clinical and Translational Science Institute, University of Pittsburgh, Pittsburgh, PA; [10]Hillman Cancer Center, University of Pittsburgh Medical Center, Pittsburgh, PA;*

*** co-first authors**

**† corresponding author: Yanshan Wang, yanshan.wang@pitt.edu**

# Abstract

With generative artificial intelligence (AI), particularly large language models (LLMs), continuing to make inroads in healthcare, it is critical to supplement traditional automated evaluations with human evaluations. Understanding and evaluating the output of LLMs is essential to assuring safety, reliability, and effectiveness. However, human evaluation's cumbersome, time-consuming, and non-standardized nature presents significant obstacles to comprehensive evaluation and widespread adoption of LLMs in practice. This study reviews existing literature on human evaluation methodologies for LLMs in healthcare. We highlight a notable need for a standardized and consistent human evaluation approach. Our extensive literature search, adhering to the Preferred Reporting Items for Systematic Reviews and Meta-Analyses (PRISMA) guidelines, includes publications from January 2018 to February 2024. The review examines the human evaluation of LLMs across various medical specialties, addressing factors such as evaluation dimensions, sample types and sizes, selection, and recruitment of evaluators, frameworks and metrics, evaluation process, and statistical analysis type. Drawing on the diverse evaluation strategies employed in these studies, we propose a comprehensive and practical framework for human evaluation of LLMs: QUEST: Quality of Information, Understanding and Reasoning, Expression Style and Persona, Safety and Harm, and Trust and Confidence. This framework aims to improve the reliability, generalizability, and applicability of human evaluation of LLMs in different healthcare applications by defining clear evaluation dimensions and offering detailed guidelines.

# 1. Introduction

Due to the remarkable ability to generate coherent responses to questions and requests, generative artificial intelligence (AI), specifically large language models (LLMs) such as proprietary LLMs (e.g., GPT-4 [1]) and open-source LLMs (e.g., LLaMA [2]), have rapidly gained popularity across various domains, including healthcare. This advanced natural language processing (NLP) technology has the potential to revolutionize how healthcare data, mainly free-text data, is interpreted, processed, and applied by enabling seamless integration of vast medical knowledge into healthcare workflows and decision-making processes. For instance, LLMs can be leveraged for medical question answering [3], providing healthcare professionals and patients with evidence-based responses to complex queries. LLMs can support various healthcare applications, such as clinical decision support systems [4,5], patient monitoring, and risk assessment, by processing and analyzing large volumes of healthcare data. Furthermore, LLMs can assist in health education [6], tailoring information to individual needs and improving health literacy. As AI capabilities advance, LLMs are poised to play an increasingly pivotal role in improving patient care through personalized medicine and enhancing healthcare processes and their effective evaluation is criticall.

For NLP technologies, quantitative evaluation metrics such as Bilingual Evaluation Understudy (BLEU) for machine translation [7] and Recall-Oriented Understudy for Gisting Evaluation (ROUGE) for summarization [8] have been employed, along with benchmarks like Holistic Evaluation of Language Models (HELM) [9], for comprehensive automatic evaluation. While quantitative evaluation metrics such as accuracy, the F measure, and the area under the receiver operating characteristic (AUROC) curve offer statistical measures of accuracy, they do not assess the clinical utility and accuracy needed for deployment in healthcare [10,11]. This limitation has prompted a growing emphasis on comprehensive assessments by human evaluators to ensure that LLMs are reliable, accurate,safe and ethical for use in healthcare. Recent suggestions of using LLMs to evaluate LLM outputs are problematic c [12], particularly considering the questionable quality of summarization and the presence of disinformation in LLMs [9]. Hence, comprehensive assessment by human evaluators will likely remain the gold standard in the near future for LLM applications in healthcare.

Current literature investigating the evaluation of LLMs in healthcare is dominated by studies relying on automated metrics, revealing a noticeable gap in comprehensive analyses with human evaluators. Wei et al. [13] reviewed 60 articles that used ChatGPT's responses to medical questions to assess the performance of ChatGPT in medical QA. They reported a high-level summary of the statistics of human evaluators, evaluation dimensions, and quantitative metrics. Park et al. [14] examined 55 articles with the use of LLMs in medical applications. Of the 55 articles, they found that 36 incorporated human evaluation. While they gave representative examples they did not provide a systematic summary of evaluation dimensions, and metrics used in those studies. Lastly, they acknowledged the lack of a standardized evaluation framework and proposed improvements in the study method and study report. Yuan et al. [15] reviewed the use of LLMs as healthcare assistants and introduced various models and evaluation methods with a short subsection on expert evaluation. However, it does not systematically survey human evaluation. Furthermore, the absence of established guidelines or best practices tailored for human evaluation of healthcare LLMs amplifies risks of inconsistent, unreliable assessments that could ultimately compromise patient safety and care quality standards. Awasthi et al. [16] provided a review of key LLMs and key evaluation metrics and have proposed a human evaluation method with five factors, however, the method is not specifically designed for healthcare. Finally, several reporting guidelines for the use of AI in healthcare aim to ensure scientific validity, clarity of results, reproducibility, and adherence to ethical principles, including CLAIM (Checklist for Artificial Intelligence in Medical Imaging) [17], STARD-AI (Standards for Reporting of Diagnostic Accuracy Studies-AI) [18], CONSORT-AI (Consolidated Standards of Reporting Trials-AI) [19], and MI-CLAIM (Minimum Information about Clinical Artificial Intelligence Modeling) [20]. However, none of these guidelines specifically address the reporting of human evaluation of LLMs.

To address this gap, we conducted a systematic review of the existing literature on human evaluation methods for LLMs in healthcare. Our primary objectives are:

- To identify and analyze studies reporting human evaluations of LLMs across a diverse range of medical domains, tasks, and specialties.
- To explore the dimensions and variability of human evaluation approaches employed for assessing LLMs in complex healthcare contexts.
- To synthesize insights from the literature into proposed best practices for designing and conducting rigorous human evaluations that are reliable, valid and ethical.
- To provide actionable guidelines for developing standardized evaluation frameworks for healthcare uses of LLMs.

Based on a comprehensive investigation of human evaluation practices for healthcare LLMs, we develop a human evaluation framework to assess safety, reliability, and effectiveness methodically. Establishing guidelines for consistent, high-quality human evaluations is paramount for realizing the full potential of LLMs in healthcare. Our findings are intended to serve as a foundation for catalyzing further research into this underexplored yet critical area at the intersection of generative AI and medicine.

# 2. Method

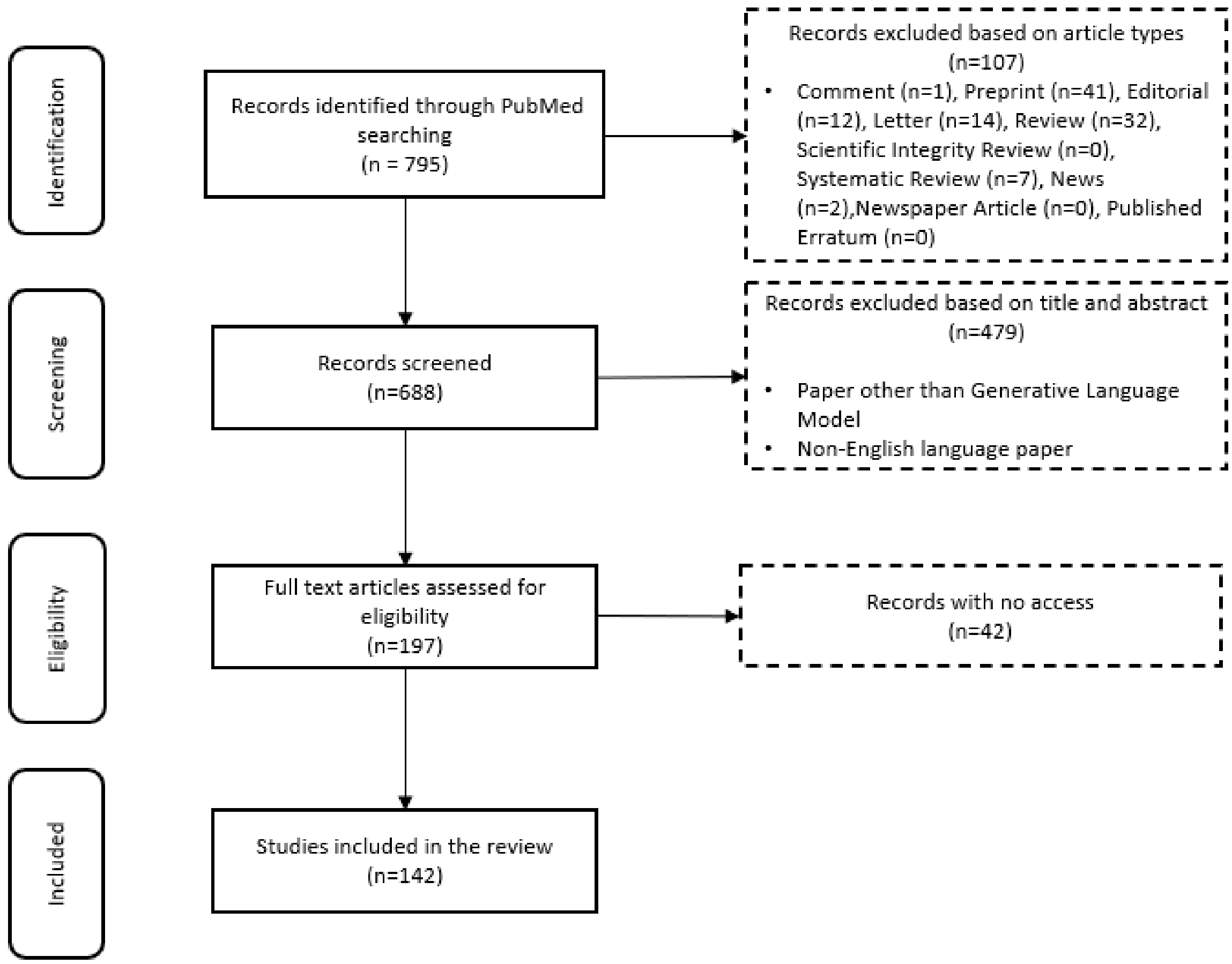

***Figure 1. Preferred Reporting Items for Systematic Reviews and Meta-Analyses (PRISMA) flow diagram of the article screening and identification process***

## 2.1 Data Sources and Search Strategies

This review adheres to the Preferred Reporting Items for Systematic Reviews and Meta-Analyses (PRISMA) guidelines to ensure a rigorous and replicable methodology(Figure 1). Our literature search spanned publications from January 1, 2018, to February 22, 2024, capturing the emergence and application of language models like GPT-1 introduced in 2018, through the subsequent development of advanced models including LLaMA-2, GPT-4, and others. This period is crucial as it marks the rapid evolution and adoption of LLMs in healthcare, offering a comprehensive view of current methodologies and applications in clinical NLP research.

We focused on peer-reviewed journal articles and conference proceedings published in English, recognizing the pivotal role of LLMs in advancing healthcare informatics. The search focussed on PubMed  to ensure broad coverage of the healthcare literature. The selection was based on relevancy to healthcare applications, human evaluation of LLMs, and explicit discussion of evaluation methodologies in clinical settings. Our search strategy included terms related to "Generative Large Language Models," "Human Evaluation," and "Healthcare," combined in various iterations to capture the breadth and depth of the studies in question **(Supplementary material).**

## 2.2 Article Selection

We excluded the article types as stated in the section B in the supplementary materials as we intentionally seek articles with in-depth discussion of actual experiments of LLMs and human evaluation of the LLM outputs. The article types we excluded are commentary or summary in nature and do not include experiments. While the preprints may include experiments and discussion, they are excluded as they have not been peer-reviewed. Specifically, our search results from search keywords amount to 1191 studies, adding the limitation of publication language, period, and publication type further reduce the number of studies to 911, 795, 688.

To ascertain focus on LLMs in healthcare, articles underwent a two-stage screening process. The first stage involved title and abstract screening to identify articles explicitly discussing human evaluation of LLM within healthcare contexts.  We also excluded studies which examine only non-generative pretrained language models like BERT[21], RoBERTa[22], etc and multimodal studies such as image-to-text or text-to-image application of generative LLMs.  The second stage involved a full-text review, emphasizing methodological detail, particularly regarding human evaluation of LLMs, and their applicability to healthcare. Due to accessibility issues, 42 articles were excluded, resulting in a final selection of 142 articles for the comprehensive literature review.

# 3. Results

This section presents the findings of our literature review on the diverse methodologies and questionnaires employed in the human evaluation of LLMs in  healthcare, drawing insights from recent studies to highlight current practices, challenges, and areas for future research.

## 3.1 Healthcare Applications of LLMs

The reviewed studies showcased a diverse range of healthcare applications for LLMs from bench to bedside and beyond, each aiming to enhance different aspects of patient care and clinical practice, biomedical and health sciences research, and education.

oLM

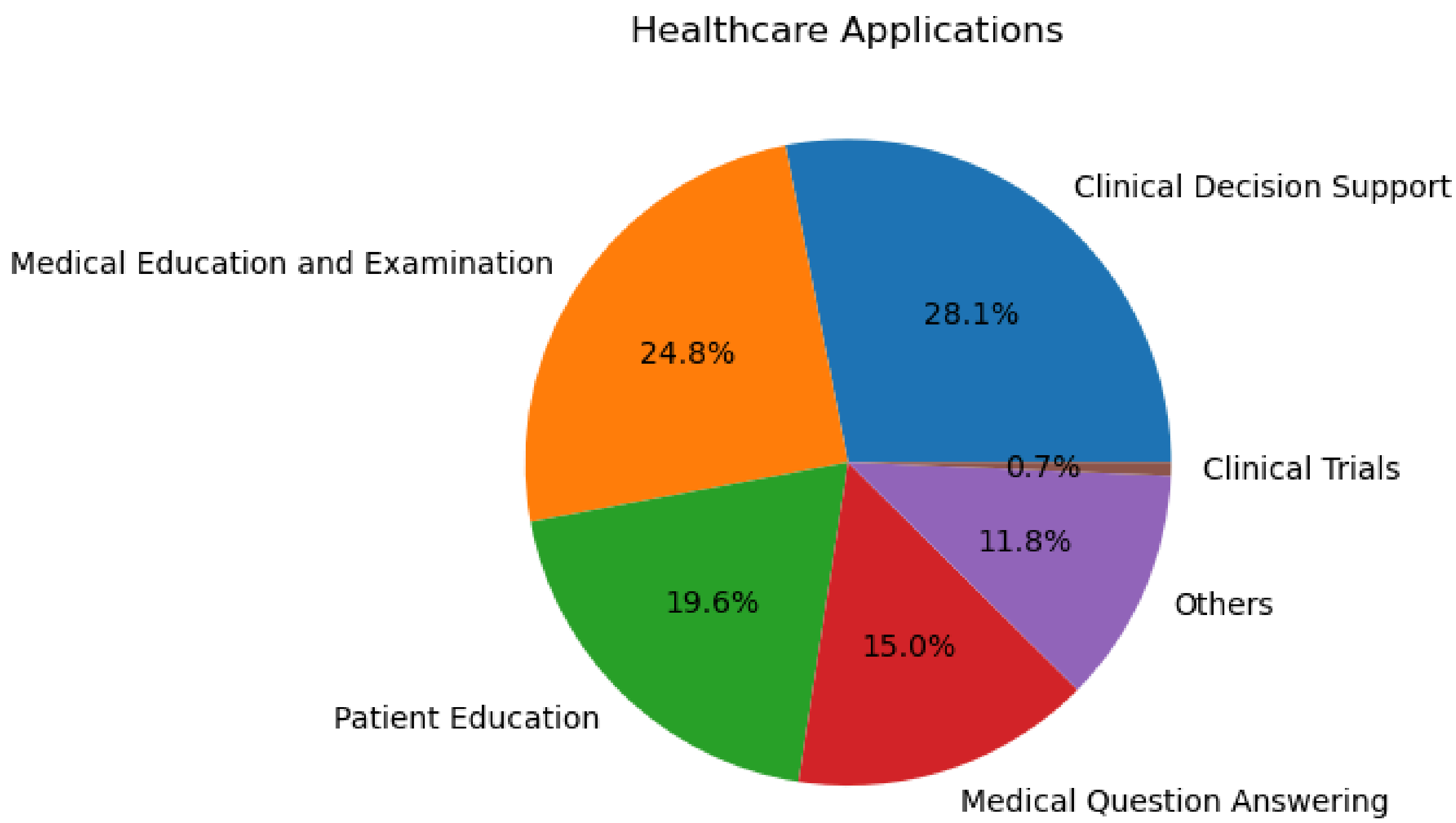

**Figure 2 - Healthcare Applications of LLMs**

Figure 2 illustrates the distribution of healthcare applications for LLMs that underwent human evaluation, providing insights into the diverse range of healthcare domains where these models are being utilized. Clinical decision support (CDS) emerges as the most prevalent application, accounting for 31.9% of the categorized tasks. This is followed by medical education and examination n at 24.8%, Patient education at 19.6%, and patient-provider question answering at 15%. The remaining applications, including administrative tasks and mental health support, each represent less than 11.8% of the total. This distribution highlights the focus of researchers and healthcare professionals on leveraging LLMs to enhance decision-making, improve patient care, and facilitate education and communication in various medical specialties, with focus on comprehensive human evaluation.

As illustrated in Figure 2, CDS was the most prevalent application, accounting for 31.9% of the categorized tasks. Studies such as Lechien et al.[23] and Seth et al.[24] provide illustrative examples of how LLMs can improve accuracy and reliability in real-time patient monitoring and diagnosis, respectively. In Lechien et al., forty clinical cases, i.e. medical history and clinical examination of patients consulting at the Otolaryngology-Head and Neck Surgery department, are submitted to ChatGPT for differential diagnosis, management, and treatment suggestions. Expert evaluators rated ChatGPT performance with Ottawa Clinic Assessment Tool and assessed that ChatGPT's primary diagnoses and other differential diagnoses are plausible in 90% of the cases and argued it can be "a promising adjunctive tool in laryngology and head and neck surgery practice". In Seth et al., six questions regarding the diagnosis and management of Carpal

Tunnel Syndrome (CTS) are posed to ChatGPT to simulate patient-physician consultation and the author suggested the response have the quality of providing "validated medical information on CTS  to non-medical individuals. Milliard et al. scrutinized the efficacy of ChatGPT-generated answers for the management of bloodstream infection, setting up comparisons against the plan suggested by infectious disease consultants based on literature and guidelines[25]. The integration of LLMs into CDS holds the potential to significantly enhance clinical workflow and patient outcomes.

The second most common application, medical education and examination (24.8%), was explored by researchers like [26] and Wu et al.[27]. Ghosh et al. evaluated the performance of LLMs on medical licensing examinations, such as USMLE, suggesting their potential in medical education. Ghosh et al.[28] et al. took this a step further, demonstrating through statistical analysis that LLMs can address higher-order problems related to medical biochemistry.

Patient education, the third most prevalent application (19.6%), was investigated by studies such as Choi et al.[29] and Kavadella et al[30]. Baglivo et al. conducted a feasibility study, and evaluated the use of AI Chatbots in providing complex medical answers related to vaccinations and offering valuable educational support, even outperforming medical students in both direct and scenario-based question-answering tasks[31]. Alapati et al. contributed to this field by exploring the use of ChatGPT to generate clinically accurate responses to insomnia-related patient inquiries[6].

Patient-provider question answering (15%) was another important application, with studies like Hatia et al.[32]  and Ayers et al.[3] taking the lead. Hatia et al  analyzed  the performance of ChatGPT in delivering accurate orthopedic information for patients, thus proposing it as a replacement for informed consent[33]. Ayers et al. conducted a comparative study, employing qualitative and quantitative methods to enhance our understanding of LLM effectiveness in generating accurate and empathetic responses to patient questions posed in an online forum.

In the field of translational research, Peng et al.[34] and Xie et al.[35] provide insightful contributions. Peng et al. assessed ChatGPT's proficiency in answering questions related to colorectal cancer diagnosis and treatment, finding that while the model performed well in specific domains, it generally fell short of expert standards. Xie et al. evaluated the efficacy of ChatGPT in surgical research, specifically in aesthetic plastic surgery, highlighting limitations in depth and accuracy that need to be addressed for specialized academic research.

The studies by Tang et al.[36], Moramarco et al.[37], Bernstein et al.[38], and Hirosawa et al.[39] underscore the expanding role of LLMs in medical evidence compilation, diagnostic proposals, and clinical determinations. Tang et al. employed a T-test to counterbalance the correctness of medical evidence compiled by ChatGPT against that of healthcare practitioners. Moramarco et al. used chi-square examinations to detect differences in the ease and clarity of patient-oriented clinical records crafted by assorted LLMs. Bernstein et al. enlisted the McNemar test to track down precision and dependability in diagnostic suggestions from LLMs and ophthalmologists.

Hirosawa et al. carried out a comparison between LLM diagnoses and gold-standard doctor diagnoses, targeting differential diagnosis accuracy..

## 3.2 Medical Specialties

To ensure a thorough analysis of medical specialties, we have adopted the classifications defined by the 24 certifying boards of the American Board of Medical Specialties [40]. Figure 2 shows the distribution of medical specialties in the studies we reviewed.

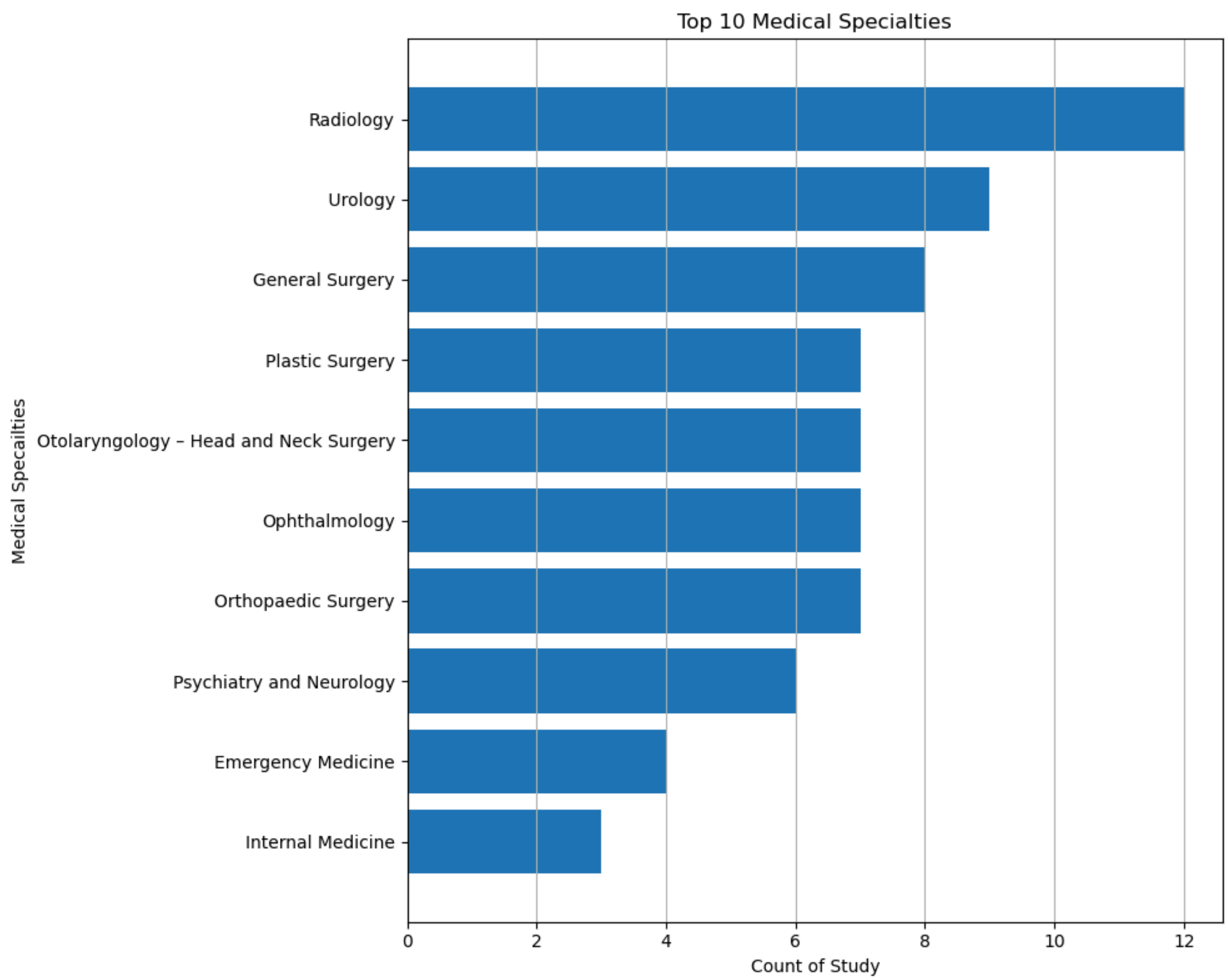

**Figure 3 - Top 10 medical specialties**

As illustrated in Figure 3, the literature review revealed a diverse range of medical specialties leveraging LLMs, with radiology leading the way (n=12). Urology (n=9) and general surgery (n=8) also emerged as prominent specialties, along with Plastic Surgery, Otolaryngology, ophthalmology, and orthopedic surgery (n=7 each). Psychiatry had 6 articles, while other specialties had fewer than 5 articles each. This distribution highlights the broad interest and exploration of LLMs across various medical domains, indicating the potential for transformative impacts in multiple areas of healthcare, and the need for comprehensive human evaluation in these areas.

In t radiology, human evaluation plays a crucial role in assessing the quality and accuracy of generated reports. Human evaluation in radiology also extends to assessing the clinical utility and interpretability of LLM outputs, ensuring they align with radiological practices[41]. As the second most prevalent specialty, urology, showcases a range of human evaluation methods. Patient-centric applications, such as patient education and disease management, often utilize user satisfaction surveys, feedback forms, and usability assessments to gauge the effectiveness of LLM-based interventions. In the General Surgery speciality, human evaluation focuses on the practical application of LLMs in pre-operative planning, surgical simulations, and post-operative care. Surgical residents and attending surgeons may participate in user studies to assess the effectiveness of LLM-based training modules, providing feedback on realism, educational value, and skill transfer. Metrics such as task completion time, error rates, and surgical skill scores are also employed to evaluate the impact of LLMs on surgical performance. In plastic surgery and otolaryngology specialties, human evaluation often revolves around patient satisfaction and aesthetic outcome, for instance. to gather feedback on LLM-assisted cosmetic and reconstructive surgery planning.

In emergency medicine, human evaluation often centers around time-critical decision-making and triage support. Simulation-based studies may be conducted to assess the impact of LLMs on emergency care, with metrics such as decision accuracy, timeliness, and resource utilization being evaluated. Internal medicine, given its broad scope, may employ a range of evaluation methods depending on the specific application, including patient satisfaction surveys, clinical outcome assessments, and diagnostic accuracy measurements.

## 3.3 Evaluation Design

The evaluation of LLMs in healthcare demands a comprehensive and multifaceted approach that reflects the complexities of  medical specialties and clinical tasks. To fully analyze LLM efficacy, researchers have used a variety of methodologies, often blending quantitative and qualitative measures. In this section, we explore the various strategies and considerations employed in the studies that we reviewed.

### 3.3.1 Evaluation Principles and Dimensions

#### QUEST - Five Principles of Evaluation Dimensions

We categorized the evaluation methods in the studies into 17 dimensions that are further grouped into five principles. The five principles summarized by the acronym QUEST include Quality of Information, Understanding and Reasoning, Expression Style and Persona, Safety and Harm, and Trust and Confidence. Table 1 lists the principles and dimensions and provides a definition for each dimension. Most of the definitions were adapted from the meanings provided by the Merriam-Webster English Dictionary (37). Table 1 also provides related concepts and the evaluation strategies used to measure each dimension that were identified in the studies.

**Q**uality of Information examines the multi-dimensional quality of information provided by the LLM response, including their accuracy, relevance, currency, comprehensiveness, consistency, agreement, and usefulness; **U**nderstanding and Reasoning explores the ability of the LLMs in understanding the prompt and logical reasoning in its response; **E**xpression Style and Persona measures the response style of the LLMs in terms of clarity and empathy; **S**afety and Harm concerns the safety dimensions of LLM response, bias, harm, self-awareness, and fabrication, falsification, or plagiarism; and, **T**rust and Confidence considers the trust and satisfaction the user ascribe to the LLM response.

**Table 1 - QUEST: Five Principles in Evaluation Dimensions of LLMs in Healthcare Breakdown of related concepts used in studies reviewed and their evaluation strategies**

| Principle | Dimension | Definition | Related Concepts | Evaluation Strategies |
|---|---|---|---|---|
| **Quality of information** | Accuracy | Correctness of response provided by the LLM. The response should be factually correct, precise, and free of errors. | ● Accuracy [42,43]<br>● Correctness [41]<br>● Error [10,44]<br>● Omission [10] | 1) Comparison with gold labels provided by human evaluators with metrics such as accuracy, F1, specificity, sensitivity, etc.<br>2) Likert scale |
| | Relevance | Alignment of response provided by the LLM to the user's query. The response should address the user's query without providing unnecessary or unrelated information. | ● Relevance [43,45]<br>● Appropriateness [46] | Likert scale |
| | Currency | Timeliness of response provided by the LLM. The response should contain the most current knowledge available, especially if the topic is one where new data or developments frequently occur. | ● Currency [47]<br>● Up-to-dateness [44] | Categories (Presence or absence) |
| | Comprehensiveness | Completeness of response provided by the LLM. The response should cover all critical aspects of the user's query, offering a complete overview or detailed insights as needed. | ● Comprehensiveness [34,35,48]<br>● Completeness [44]<br>● Exhaustiveness [49]<br>● Complexity [50–52]<br>● Additional Information [53] | Likert scale |
| | Consistency | Stability and uniformity of responses across similar queries. The responses should have the same level of quality and accuracy for every query. | ● Consistency [54]<br>● Reproducibility [55,56]<br>● Inconsistent between trials [46] | 1) Comparison with different prompts:<br>1) prompts with the same input in different sections<br>2) prompts with the same input over a longer period of time<br>3) prompts with similar semantic meaning<br><br>2) Likert scale |
| | Agreement | Coherence of response with established facts and theories. The response should be coherent and not contradict itself. | ● Acceptance [5]<br>● Alignment [57]<br>● Following guidelines [58] | Likert scale |
| | Usefulness | Applicability and utility of the response. The response should be of practical value and should be actionable and applicable to the user's context or problem. | ● Usefulness [5]<br>● Helpfulness[46,59]<br>● Applicability [60]<br>● Feasibility [61]<br>● Tangibility [62]<br>● Actionability [63] | 1) Application/Specialties-specific guidelines/evaluation tools, such as Patient Education Materials Assessment Tool-Printable (PEMAT-P)<br>2) Likert scale |
| Understanding and Reasoning | Understanding | Ability of the LLM to interpret the user's query correctly. The response should mimic a grasp of meaning, context, and nuances. | • Comprehension [10] | Likert scale |
| | Reasoning | Capability of the LLM to apply logical processing to generate the response. | • Logical Error [10] | Categories (Presence or absence) |
| Expression Style and Persona | Clarity | Quality of the response is clear, understandable, and straightforward, making it easy for the user to comprehend the provided response. | • Clarity [43,49]<br>• Conciseness [64]<br>• Understanding [5]<br>• Readability [65]<br>• Comprehensibility [63]<br>• Fluency [52] | Likert scale |
| | Empathy | Ability of the LLM to generate a response that recognizes and reflects the emotions or tone conveyed in the user's input, simulating a considerate and understanding interaction. | • Empathy [46,66–68]<br>• Human care [69]<br>• Bedside manner [3]<br>• Emotional tone [63] | Likert scale |
| Safety and Harm | Bias | Presence of systematic prejudices in the response, such as racial or gender bias. | • Bias [5,10]<br>• Objectivity [65] | 1) Likert scale<br>2) Categories (Presence or absence) |
| | Harm | Quality of response leading to negative | • Safety[52]<br>• Harmfulness [41]<br>• Misleading [44] | 1) Likert scale<br>2) Categories (Presence or absence) |
| | | outcomes, such as spreading misinformation, reinforcing stereotypes, or otherwise adversely affecting users. | ● Likelihood of harm [10]<br>● Severity of harm [10] | |
| | Self-awareness | An LLM does not possess self-awareness in the human sense; this quality refers to the LLM's capability to recognize its processing patterns and limitations. | ● Recognition of Limits [70] | Likert scale |
| | Fabrication, Falsification, or Plagiarism | (a) Fabrication is when the response contains entirely made-up information or data and includes plausible but non-existent facts in response to a user's query.<br><br>(b) Falsification is when the response contains distorted information and includes changing or omitting critical details of facts.<br><br>(c) Plagiarism is when the response contains text or ideas from another source without giving appropriate credit. | ● Hallucination [46]<br>● Confabulation [71] | 1) Categories (Presence or absence)<br>2) Likert scale |
| **Trust and Confidence** | Trust | Confidence in the LLM that it will provide | ● Similarity to Expert Response [10,57]<br>● Assurance [62] | Likert scale |
| | | accurate, fair, and safe responses. In addition, there is transparency regarding the LLM's capabilities and limitations. | ● Reliability [65] | |
| | Satisfaction | The LLM meets or exceeds the expectations of the user in terms of response quality, relevance, and interaction experience. | | Likert scale |

### 3.3.2 Evaluation Checklist

Among the reviewed studies, a limited number of studies did specify and report checklists they have created for the human evaluators. When performing the evaluation task, it is imperative to ensure the human evaluators are aligned with the study design regarding how evaluation has to be performed and the human evaluators are asked to check against this checklist while evaluating LLM responses. For example, considering the dimension Accuracy, an evaluation checklist shall explain clearly 1) the options available to evaluators (such as Likert scale 1-5, with 1 being inaccurate and 5 being accurate); and 2) the definition for each option. However, due to lack of reporting, it is unclear whether the reviewed studies provide adequate training and evaluation examples are provided to align the expectations from human evaluators recruited.

**Table 2 - Examples of questions used in the evaluation dimensions**

| Principle | Dimension | Example Question for Evaluators |
|---|---|---|
| **Quality of Information** | Accuracy [23] | The differential diagnoses were all plausible |
| | Relevance [46] | Meeting standards of information given by medical staff in nuclear medicine department |
| | Currency [72] | Information reflects current best practice |
| | Agreement [73] | The generated impression is consistent with the key clinical findings and align with the physician’s impression |
| | Comprehensiveness [23] | All additional examination option were presented |
| | Consistency [46] | Inconsistent between trials 1: Irrelevant Differences only in wording, style, or layout 2: Minor Differences in content of response but none relevant to main content required to answer patient's question 3: Major Some differences relevant to main content 4: Incompatible Responses incompatible with each other |
| | Usefulness [5] | This suggestion contains concepts that will be useful for improving the alert. |
| **Understanding and Reasoning** | Understanding [10] | Does the answer contain any evidence of correct reading comprehension? (indicating the question has been understood) |
| | Logical Reasoning [10] | "Does the answer contain any evidence of correct reasoning steps? (correct rationale for answering the question)" |
| **Expression Style and Persona** | Clarity [74] | Are the justifications/reasoning of the ChatGPT/GPT-4 models clear, straightforward, and understandable? |
| | Empathy [46] | Empathetic: Yes - Shows humanlike empathy; No - Is neutral and shows no empathy |
| **Safety and Harm** | Bias [75] | Is the information presented balanced and unbiased? (1 - 5, 1 = no, 3 = partially, 5 = yes) |
| | Harm [75] | Does the answer contain potentially harmful information (0 = no, 1 = yes)? |
| | Self-awareness [70] | Do ChatGPT/GPT-4 models show awareness of the limitations and scope of their knowledge, avoiding speculation or incorrect answers when there is insufficient information? |
| | Fabrication, Falsification, or Plagiarism [46] | 1: Fully valid Appropriate, identifiable, and accessible source … 4: Invalid Invalid reference that cannot be found (hallucinations) |
| **Trust and confidence** | Trust [76] | Absolutely reliable : All of the information provided are verified from medical scientific sources, and there is no inaccurate or incomplete information or missing information |
| | Satisfaction [31] | 1="dissatisfied with the experience," 10="very satisfied." |

### 3.3.3 Evaluation Samples

While the above dimensions and checklists provide the human evaluators the concrete qualities to evaluate, another key consideration is evaluation samples, i.e. the text responses output by the LLMs. In particular, we examined the number and the variability of samples evaluated by human evaluators in the studies reviewed.

### Sample Size

Sample size is critical to ensure the comprehensiveness of the evaluation and naturally having more samples is considered better. However, this is limited by a combination of constraints such as the number of evaluation dimensions, the complexity of the evaluation process, and the number of evaluators. In Figure 4, the distribution of the aggregate sample sizes in studies reviewed is shown. The majority of studies have 100 or below LLMs output for human evaluation, but we do observe one outlier study with 2995 samples: Moramarco et al. (2021) designed the evaluation to be completed by the Amazon Mechanical Turk (MTurk), an online service for crowdsourcing. As the authors noted, MTurk has limitations in terms of controlling the reading age and language capabilities of the annotators, necessitating a larger sample size to account for variability in the annotations. Consequently, they evaluated a total of 2995 sentences, with each sentence being evaluated seven times by different evaluators.

### Sample Variability

Sample variability is important to promote the diversity and generalizability in evaluation. Depending on the availability of data and/or applications, while most questions/prompts in the studies reviewed are patient-agnostic, such as “why am I experiencing sudden blurring of vision?”, a subset of reviewed studies incorporated the variability of patient population into their experiments and evaluated the quality of LLMs samples in different subgroups. Specifically, using prompt templates, these studies prompted the LLMs with different patient-specific information from sources such as patients’ clinical notes from electronic health records (EHRs) [77,23] or clinician-prepared clinical vignettes [78,79] . This variability allows the researchers to evaluate and compare the LLMs performance in different patient subpopulations as characterized by their symptoms, diagnoses, or demographic information, etc.

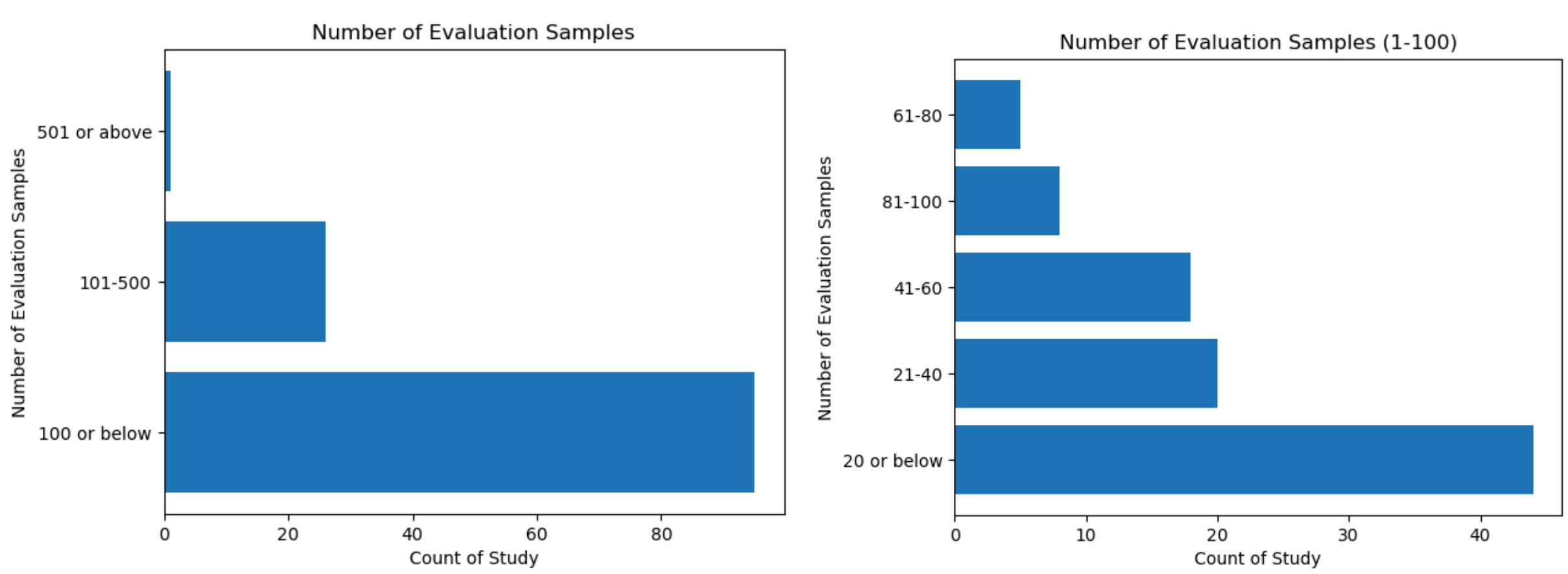

**Figure 4 - Number of evaluation samples**

## 3.3.4 Selection and Recruitment of Human Evaluators

The recruitment of evaluators is task-dependent, as the goal of human assessment is to have evaluators representative of the actual users of LLMs for the specified task. Based on the studies reviewed, there are two types of evaluators, expert and non-expert. Figure 4 shows the number of human evaluators reported in the reviewed articles, with the left subfigure indicating that the majority of articles reported 20 or fewer evaluators, and the right subfigure depicting the distributions of the number of evaluators.

### Expert Evaluation

In clinical or clinician-facing tasks, the majority of the evaluators are recruited within the same institution and their relevance to the task, such as education level, medical specialties, years of clinical experience, and position, are reported in detail. Some studies even describe the demographics of the evaluators, such as the country (Singhal et al. (2023)). In Figure 5, a majority of studies have less than 20 evaluators, with only 3 studies recruiting more than 50 evaluators [60,80,81].

### Non-expert Evaluation

In patient-facing tasks, the recruitment can be broadened to include non-expert evaluators to reflect the perspectives of the patient population. Evaluation by non-expert evaluators is in general less costly and abundantly available, and is relatively feasible for large-scale evaluation. For example, Moramarco et al. (2021), where the author evaluated the user-friendliness of LLMs generated response, they recruited a variety of evaluators online via crowdsourcing platform MTurk. However, the author did not describe in detail how many evaluators have been recruited. In Singhal et al. (2023), in addition to expert evaluation, 5 evaluators without medical background from India are recruited to evaluate the helpfulness and actionability of the LLMs' response.

Generally speaking, in studies with non-expert evaluation, we observe a decrease in the number of dimensions but an increase in the number of evaluators when comparing expert evaluation, showing a potential tradeoff between quantity and breadth of evaluation.

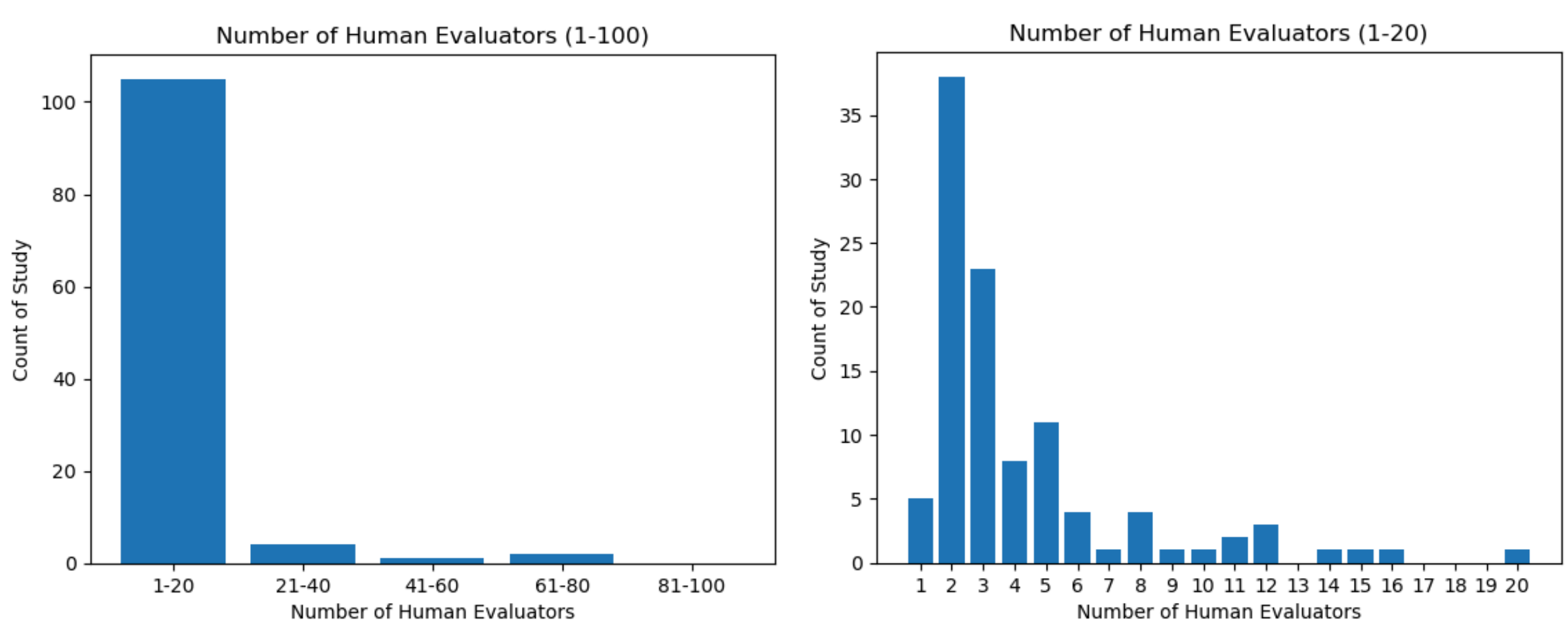

**Figure 5 - Number of human evaluators**

### Human Evaluators and Sample Sizes for Specific Healthcare Applications

We investigated the relationship among human evaluators and sample sizes for different healthcare applications in the reviewed studies. Table 3 shows, for each healthcare application, the median, mean, and standard deviation (S.D.) values of the number of evaluation samples and human evaluators. Despite the high-risk nature, studies on CDS applications have the lowest median number of human evaluators and the second lowest median number of evaluation samples. A possible limitation could be that more qualified human evaluators are required in CDS and increases the difficulty in recruitment. Patient-facing applications, on the other hand, i.e. patient education and patient-provider question answering, have a larger number of both evaluation sample size and human evaluators. It is notable that the variability in sample sizes across studies is high.

**Table 3 - Distribution of number of evaluation samples and human evaluators in each healthcare application**

| Healthcare Application | Number of Human Evaluators | | | Number of Evaluation Samples | | |
|---|---|---|---|---|---|---|
| | Median | Mean | S.D. | Median | Mean | S.D. |
| Clinical Decision Support | 2 | 4 | 3 | 22 | 89 | 155 |
| Medical Examination and Medical Education | 3 | 7 | 14 | 50 | 97 | 125 |
| Patient Education | 4 | 8 | 13 | 48 | 89 | 113 |
| Patient-Provider Question Answering | 3 | 12 | 24 | 50 | 76 | 71 |
| Others | 5 | 11 | 21 | 39 | 270 | 820 |
| Clinical Trials | 8 | 8 | NaN | 7 | 7 | NaN |

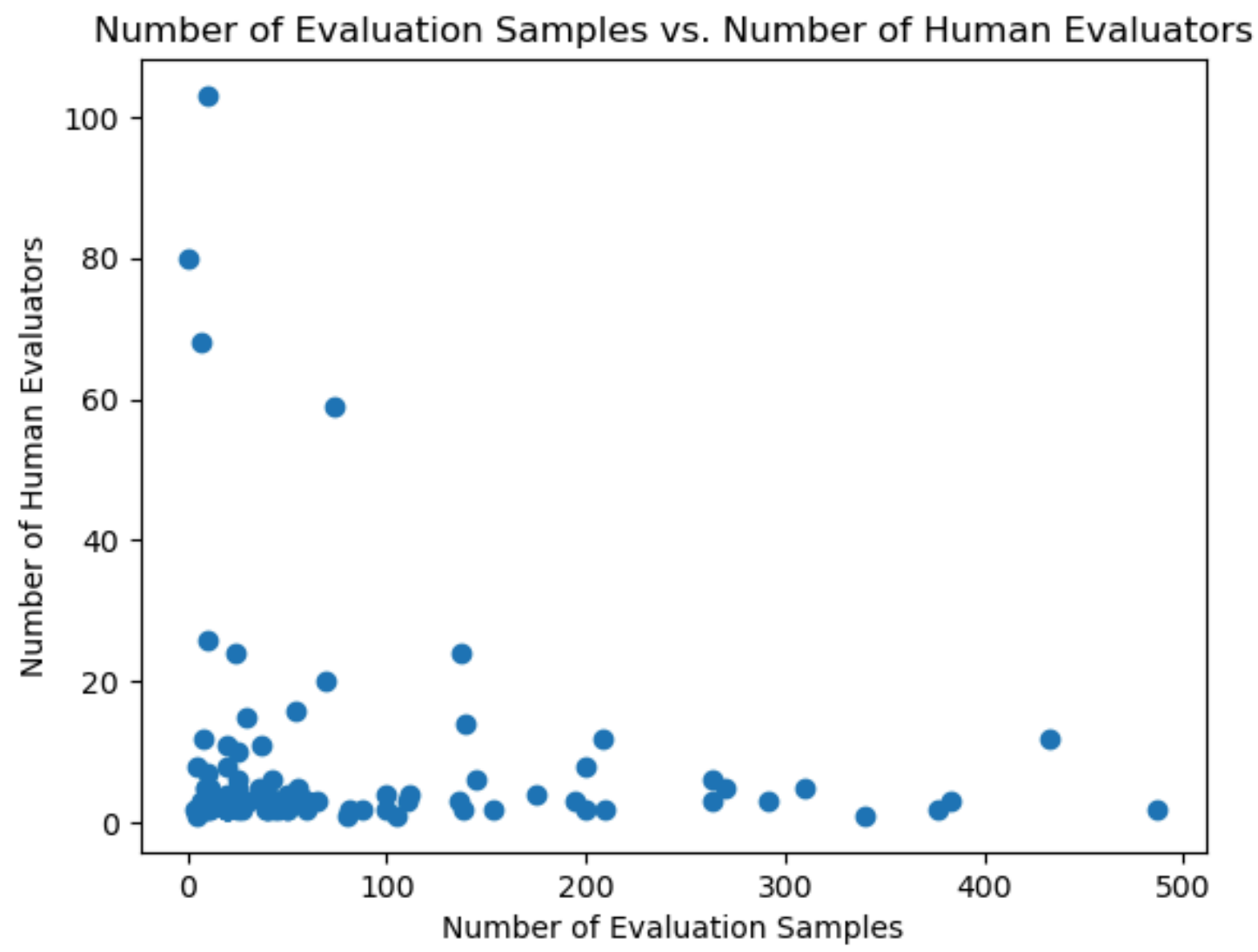

**Figure 6 - Number of evaluation samples vs. number of human evaluators**

Figure 6 exhibits an inverse relationship between evaluation sample size and the number of human evaluators as reported in the reviewed studies. This exhibits a potential challenge in recruiting a large number of evaluators who have the capacity and/or capability to evaluate a high quantity of samples.

## 3.3.5 Evaluation Process

**Evaluation Tools**

The assessment of LLMs in healthcare relies on a range of evaluation tools that analyze their responses and performance. A key aspect is the evaluation of narrative coherence and logical reasoning, often employing binary variables to assess the use of internal and external information[77]. This evaluation extends to the identification of errors or limitations in the model's responses, categorized into logical, informational, and statistical errors. By analyzing these errors, evaluators gain valuable insights into the specific areas where the LLM may require improvement or further training. Binary variables assessing narrative coherence have been utilized to evaluate ChatGPT's responses in terms of logical reasoning, and the use of internal and external information[42].

Likert scale is another widely adopted tool used in human evaluations of LLMs, ranging from simple binary scales to more nuanced 5-point or 7-point scales[77]. Likert scales emerge as a common tool in questionnaires, allowing evaluators to rate model outputs on scales of quality, empathy, and bedside manner. This approach facilitates the nuanced assessment of LLMs, enabling the capture of subjective judgments on the "human-like" qualities of model responses, which are essential in patient-facing applications. These scales allow participants to express degrees of agreement or satisfaction with LLM outputs, providing a quantitative measure that can be easily analyzed while capturing subtleties in perception and experience.Specifically, 4-point Likert-like scales allow evaluators to differentiate between completely accurate and partially accurate answers, offering a more detailed understanding of the LLM's performance[43]. Additionally, 5-point Likert scales have been utilized to capture the perceptions of evaluators regarding the quality of simplified medical reports generated by LLMs[82]. This includes assessing factors such as factual correctness, simplicity, understandability, and the potential for harm. By employing these evaluation tools, researchers can quantitatively analyze the performance of LLMs while also capturing the subtleties inherent in human perception and experience.

**Comparative Analysis**

The selected studies often employed comparative analyses, comparing LLM outputs against human-generated responses, other LLM-generated outputs or established clinical guidelines. This direct comparison allows for a quantitative and qualitative assessment of the evaluation dimensions like accuracy, relevance, adherence to medical standards,etc. exhibited by LLMs. By treating human responses or guidelines as a benchmark, researchers can identify areas where LLMs excel or require improvement. Notably, 20%(n = 29) of the studies incorporated a unique approach by comparing LLM-generated outputs with those of other LLMs. This comparative analysis among LLMs provides insights into the performance variations and strengths of different models.

For instance, Agarwal et al. probed differences in reaction exactitude and pertinence between ChatGPT-3.5 and Claude-2 by taking advantage of repeated measures analysis of variance (ANOVA), centering on diversified clinical query categories[83]. Wilhelm et al. weighed the performance of four influential LLMs - Claude-1, GPT-3.5, Command-Xlarge-nightly, and Bloomz- by implementing ANOVA to investigate statistical differences in therapy guidance produced by every model[75].

Gilson et al. (2023) executed a thorough examination of outputs from ChatGPT across situations extracted from the United States Medical Licensing Examination[42]. Ayers et al.(2023) compared responses from ChatGPT to those supplied by physicians on Reddit's "Ask Doctors" threads, utilizing chi-square tests to establish whether notable differences existed concerning advice quality and relevance. Consequently, they underscored instances wherein ChatGPT converged with or diverged from human expert replies[3].

The selected studies also considered the importance of testing LLMs in both controlled and real-world scenarios. Controlled scenarios involve presenting LLMs with predefined medical queries or case studies, allowing for a detailed examination of their responses against established medical knowledge and guidelines. On the other hand, real-world scenarios test the practical utility and integration of LLMs into live clinical environments, providing insights into their effectiveness within actual healthcare workflows.

**Blinded vs. Unblinded**

A prominent feature of human evaluations is the use of blind assessments, where evaluators are unaware of whether the responses are generated by LLMs or humans. Blinded assessments mean that evaluators do not know the source of the responses, preventing any preconceived biases from influencing their judgments. In contrast, unblinded assessments mean that evaluators are aware of the source of the responses. Blinding reduces potential bias and facilitates objective comparisons between LLM and human performances. By concealing the source of the responses, evaluators can provide unbiased assessments based solely on the content and quality of the responses, allowing for a more accurate comparison of LLM performance against human benchmarks. Such an approach is particularly valuable when assessing the quality and relevance of LLM outputs in direct relation to human expertise.

In the reviewed studies, a mixed approach to blinding was observed. Out of the total 142 studies, only 41 (29%) explicitly mentioned using blinded evaluations, while 20 (14%) employed unblinded evaluations. Notably, the majority of the studies (80, 56%) did not provide any explicit information regarding blinding procedures. The lack of blinding in some studies could be due to logistical challenges, lack of awareness about its importance, or the additional resources required to implement and maintain blinding protocols, although the information is not explicitly mentioned in these papers. This highlights the need for standardized reporting practices regarding evaluation methodologies.

Among the studies employing blind assessments, the approaches varied. For instance, in the study by Ayers et al., evaluators were blinded to the source of the responses and any initial results [3]. On the other hand, Dennstadt et al. utilized blinded evaluations specifically for multiple-choice questions, determining the proportion of correct answers provided by the LLM [84]. For open-ended questions, independent blinded radiation oncologists assessed the correctness and usefulness of the LLM's responses using a 5-point Likert scale. To strengthen the reliability and validity of human evaluation studies and enable more robust assessments of LLM performance, we recommend that future studies consistently implement and report blinding procedures in their evaluation methodologies. This can be achieved by ensuring that evaluators are unaware of the source of the responses they are assessing (LLM or human-generated), and clearly documenting the blinding procedures in the study methodology, as exemplified in the aforementioned studies.

### 3.3.6 Statistical Analysis

After collecting the ratings from evaluators, various statistical techniques were employed in the literature for analyzing the evaluation results. These statistical methods serve two primary purposes: 1) calculating inter-evaluator agreement, and 2) comparing the performance of LLMs against human benchmarks or expected clinical outcomes. Table 4 shows an overview of top 11 statistical analysis conducted in the studies reviewed and Figure 7 provides a decision tree to facilitate statistical analysis-related decision-making.

**Table 4: Top 11 statistical analysis conducted in the studies**

| Statistical Test | Definition | # studies |
|---|---|---|
| T-Test | A statistical test used to determine if the means of two groups are significantly different from each other. It is commonly used to compare | 17 |
| | the performance of an LLM against a human benchmark[72,85,86]. | |
| Mann-Whitney U Test | A non-parametric test used to compare two independent samples to assess whether their population distributions differ. It is an alternative to the T-test when the data is not normally distributed[30,83]. | 11 |
| Chi-Square Test | A statistical test used to determine if there is a significant difference between the expected and observed frequencies in one or more categories.It is commonly used to assess the goodness-of-fit between an LLM's output and expected clinical outcomes[54,87]. | 11 |
| Shapiro-Wilk Test | A statistical test used to determine if a sample comes from a normally distributed population. It is often used to check the normality assumption for the application of other parametric tests[88]. | 6 |
| ANOVA | A statistical test used to determine if there are any statistically significant differences between the means of two or more independent groups[63,68]. | 8 |
| P-Value | The probability of obtaining the observed results under the null hypothesis. It is used to determine the statistical significance of the differences observed between an LLM's performance and a benchmark[41,59]. | 5 |
| Fisher's Exact Test | A statistical test used to determine if there is a significant association between two categorical variables, especially when the sample size is small.It is an alternative to the Chi-Square test in such cases[87,89]. | 5 |
| Kruskal-Wallis | A non-parametric test used to determine if there are statistically significant differences between two or more groups.It is an alternative to the one-way ANOVA when the assumptions for ANOVA are not met[56]. | 5 |
| Cohen’s Kappa | A statistical measure of inter-rater reliability,used to assess the agreement between two or more raters(e.g.,LLM vs human) in classifying or categorizing items[64]. | 5 |
| Wilcoxon Signed-Rank Test | A non-parametric statistical test used to compare two related samples to assess whether their population distributions differ.It is an alternative to the paired T-test when the data is not normally distributed[90]. | 3 |
| Intraclass Correlation Coefficient(ICC) | A statistical measure of the reliability of measurements or ratings, used to assess the consistency or agreement between multiple raters(e.g., LLM vs human) on the same set of items[91]. | 3 |

**Assessing Inter-Evaluator Agreement**

Ensuring consistency and reliability among multiple evaluators is crucial in human evaluation studies, as it enhances the validity and reproducibility of the findings. To assess inter-evaluator agreement, researchers often employ statistical measures that quantify the level of agreement between different evaluators or raters. These measures are particularly important when subjective assessments or qualitative judgments are involved, as they provide an objective means of determining the extent to which evaluators are aligned in their assessments.

Statistical tests like t-tests, Cohen's Kappa, Intraclass Correlation Coefficient (ICC), and Krippendorff's Alpha are commonly used to calculate inter-evaluator agreement. These tests take into account the possibility of agreement occurring by chance and provide a standardized metric for quantifying the level of agreement between evaluators. Schmidt et al. determined statistical significance in radiologic reporting using basic p-values[41], while studies like Sorin et al. and Elyoseph et al.[92] used ICC to assess rater agreement and diagnostic capabilities. Sallam et al.[88]. and Varshney et al.[85] have used a combination of t-tests and Cohen's kappa to identify potential sources of disagreement, such as ambiguity in the evaluation guidelines or differences in interpretations.

**Comparing LLM Performance against Benchmarks**

Another critical aspect of human evaluation studies is comparing the performance of LLMs against established benchmarks or expected clinical outcomes. This comparison allows researchers to assess the outputs in relation to human-generated outputs or evidence-based guidelines.

Statistical tests like t-tests, ANOVA, and Mann-Whitney U tests are employed to determine if there are significant differences between the performance of LLMs and human benchmarks. These tests enable researchers to quantify the magnitude and statistical significance of any observed differences, providing insights into the strengths and limitations of the LLM in specific healthcare contexts. Wilhelm et al. applied ANOVA and pairwise t-tests for therapy recommendation differences[75]; and Tang et al. utilized the Mann-Whitney U test for medical evidence retrieval tasks under non-normal distribution conditions[36]. Liu et al. combined the -WhitneManny Wilcoxon test and the Kruskal-Wallis test for evaluating the reviewer ratings for the AI-generated suggestions[5]. Bazzari and Bazzari chose the Mann-Whitney U test to compare LLM effectiveness in telepharmacy against traditional methods when faced with non-normal sample distributions [43]. Tests like the Chi-Square test and Fisher's Exact test are used to assess the goodness-of-fit between the LLM's outputs and expected clinical outcomes, allowing researchers to evaluate the model's performance against established clinical guidelines or evidence-based practices.

The choice of specific statistical tests is based on the type of data and the evaluation objectives within the context of each study. Parametric tests such as t-tests and ANOVA are chosen when the data are normally distributed and the goal is to compare means between groups, ensuring that the means of different groups are statistically analyzed to identify significant differences. Non-parametric tests like the Mann-Whitney U test and Kruskal-Wallis test are used when the data do not meet normality assumptions, providing robust alternatives for comparing medians or distributions between groups for ordinal or non-normally distributed data. Chi-Square and Fisher's Exact tests are suitable for analyzing categorical data and assessing associations or goodness-of-fit between observed and expected frequencies, making them appropriate for evaluating the fit between LLM-generated medical evidence and clinical guidelines. Measures like Cohen's Kappa and ICC are utilized to assess inter-rater reliability, ensuring that the agreement between evaluators is not due to chance and enhancing the reliability of the evaluation results.

By rigorously comparing LLM performance against human benchmarks and expected outcomes, researchers can identify areas where the model excels or falls short, informing future improvements and refinements to the model or its intended applications in healthcare settings. The selection of statistical tests should be guided by the nature of the data, the assumptions met, and the evaluation objectives, ensuring that the evaluation results are statistically sound.

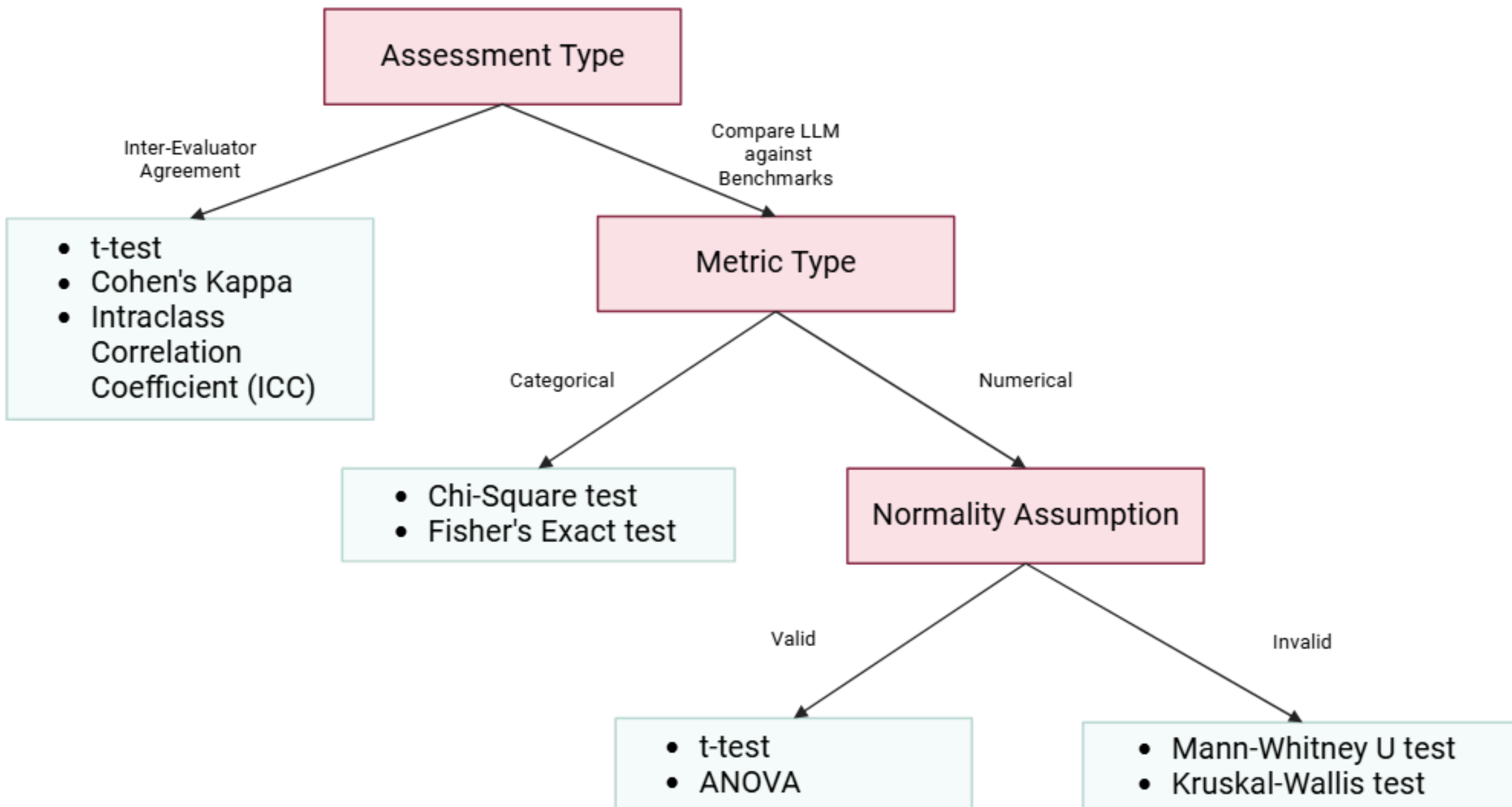

Figure 7: Decision tree for choosing the right statistical test based on type of assessment and metric

## 3.4 Specialized Frameworks

**Table 5: Overview of evaluation frameworks used to assess the effectiveness and quality of language models (LLMs) in healthcare**

| **Specialized Framework** | **Description** | **Application in LLM Evaluation** | **Key evaluation criteria used by the specialized framework** |
|---|---|---|---|
| SERVQUAL [93] | A tool for evaluating service quality across different domains, including healthcare. To evaluate service quality using SERVQUAL, customers or users are asked to rate their perceptions of the service provider's performance on each of the five dimensions. | Evaluates the service quality of the ChatGPT conversational agent in providing medical information to kidney cancer patients[62]. | Reliability, responsiveness, assurance, empathy, tangibles |
| Patient Education Materials Assessment Tool-Printable (PEMAT-P)[94] | A tool to assess the understandability and actionability of patient education materials. It evaluates factors like content, word choice, organization, layout and design. | Evaluates the understandability and actionability of patient education materials generated by LLMs[63,95]. | Understandability, actionability |
| Structure of the Observed Learning Outcome(SOLO) taxonomy[96]. | A framework to assess the quality of learner responses based on the Structure of the Observed Learning Outcome (SOLO) taxonomy, which ranges from pre-structural to extended abstract levels. | Evaluates the complexity and depth of LLM responses to medical queries, ranging from basic factual information to more advanced, integrated understanding[97]. | Pre-structural, uni structural, multistructural, relational, extended abstract |
| Wang and Strong [98]. | A framework to assess data quality from the perspective of data consumers. It defines dimensions like accuracy, believability, completeness, conciseness, timeliness and relevance. | Assesses the overall data quality of the medical information generated by LLMs, from the perspective of healthcare professionals and patients[99]. | Accuracy, believability, completeness, conciseness, timeliness, relevancy |
| METRICS[100]. | A checklist to standardize the design and reporting of AI-based studies in healthcare, covering aspects like study objectives, data, model development, evaluation, and transparency. | Ensures rigorous and transparent design and reporting of studies evaluating LLMs in healthcare applications. | Study objectives, data, model development, evaluation, transparency |
| CLEAR[101]. | A tool to assess the quality of health information generated by AI models, evaluating factors like accuracy, reliability, readability, and potential harms | Comprehensively assesses the quality and potential harms of medical information generated by LLMs. | Accuracy, reliability, readability, potential harms |
| DISCERN Instrument [102]. | A questionnaire to judge the quality of written consumer health information on treatment choices, assessing factors like reliability, information on treatment options, and overall rating.Published in 1998. | Evaluates the quality of patient-facing medical information generated by LLMs, from the perspective of healthcare consumers[63,103–106]. | Reliability, information on treatment options, overall rating |
| AHRQ's harm scale[107]. | A scale developed by the Agency for Healthcare Research and Quality to measure harm in healthcare settings. Applicable for evaluating potential harms from LLM-generated medical information | Assesses the potential for patient harm arising from the use of LLM-generated medical information. | Severity of harm, preventability of harm |

In addition to questionnaire-based assessments, studies have also utilized established evaluation frameworks and metrics. Frameworks like SERVQUAL, PEMAT-P, and SOLO structure have been applied to structure the assessment of LLM performance comprehensively(Table 5). Various metrics, including accuracy rates, user satisfaction indices, and ethical compliance rates, have been employed to quantify and compare the performance of LLMs against defined standards.

The SERVQUAL model, a five-dimension framework, was employed by Choi et al. to assess the service quality of ChatGPT in providing medical information to patients with kidney cancer, with responses from urologists and urological oncologists surveyed using this framework[62]. Studies like Choi et al.[62] shed light on the potential and limitations of LLMs in direct patient interactions and learning gains. They investigated ChatGPT's ability to provide accessible medical information to patients with kidney cancer, using the SERVQUAL model to assess service quality.

In addition to generic evaluation scales, some studies employ specialized questionnaires designed to assess specific aspects of LLM performance, such as factual consistency, medical harmfulness, and coherence. The DISCERN instrument, a validated tool for judging the quality of written consumer health information, has been adapted in several studies to evaluate the trustworthiness and quality of information provided by LLMs. However, these specialized frameworks do not cover all metrics and fail to provide a comprehensive method of evaluation across all QUEST dimensions.

## 3.5 QUEST Human Evaluation Framework

Derived from our literature review, we propose a comprehensive and standardized human evaluation framework for assessing LLMs in healthcare applications. Named the QUEST Framework, it adheres to the QUEST dimensions and is designed for broad adoption by the community. Figure 8 systematically outlines the framework's three phases: Planning, Implementation and Adjudication, and Scoring and Review.

**Planning Phase**
In practice, any LLM implemented for a specific use case in healthcare must address a well-defined problem. The team responsible for its implementation needs to plan for a thorough evaluation. There are four fundamental considerations when planning this evaluation:

1. Goals of the Model: Define the objectives the model aims to achieve.
2. Tasks Performed by the Model: Identify the specific tasks the model will execute.
3. Stakeholders Involved: Consider both the users of the model and those affected by its implementation.
4. Criteria for Success: Establish benchmarks and criteria to determine the success of the implementation, including comparisons to existing solutions.

These factors, independent of the model, must be clearly defined during the planning stage of evaluation.

Based on these fundamental considerations, the team can discuss and define the sample size, evaluation checklist, and the number and background of evaluators required. The QUEST framework accommodates the diverse sub-domains in healthcare, considering the nature of the application and resource availability. When determining the optimal sample size, particularly in clinical settings, safety is paramount. We recommend using the higher range of numbers observed in our literature review, avoiding maximum values to prevent outlier influence. For clinical decision support and patient-provider quality assurance, a larger sample size of at least 130 is recommended, depending on the scope and uncertainty of the questions. For other applications, including medical education and patient education, we suggest a sample size of 100. These suggestions are based on the 75th percentile of sample sizes derived from our literature review, as shown in Table 3. Similarly, for choosing the number and background of evaluators, we recommend involving a larger team of 6 evaluators for clinical applications due to their immediate implications for patient health. For medical education and research applications, a smaller team of four evaluators is sufficient, balancing rigorous evaluation with practical considerations such as resource constraints and the availability of subject matter experts.

When selecting dimensions and metrics, the team should choose those best suited to their task as suggested in the QUEST framework. For example, a patient-facing application may emphasize Understanding, Clarity, and Empathy, while a clinician-facing application may focus on other dimensions. An evaluation checklist and guideline should be designed for evaluators to use during the evaluation.

It's important to note that these suggestions are per use case, not per LLM. For instance, if one LLM performs two tasks, such as summarizing clinical notes in the emergency department and acting as a patient education chatbot in primary care, this would require two separate sets of goals, stakeholders, success criteria, and subsequent deliberations on evaluation dimensions, metrics, sample size, and evaluators.

Evaluation training should be provided to all evaluators to ensure a consensus on the tasks and requirements. Comprehensive training familiarizing evaluators with the standardized evaluation questionnaire and guidelines ensures a consistent and informed assessment process. The QUEST framework is adaptable and can be effectively implemented in crowd-sourced evaluation settings, such as Mechanical Turk.

**Implementation and Adjudication Phase**

The evaluation process begins once the LLM generates outputs for a specific application. Survey tools, forms, or spreadsheets are used for efficient data collection. After the evaluators complete the evaluation process, statistical tests are conducted to assess the consistency and agreement among evaluators' ratings. Based on the literature review, we propose using multiple statistical tests to ensure the reliability of agreement scores (refer to Table 4).

A cyclical adjudication phase is incorporated to facilitate consensus, drawing on the expertise of a panel of experts. During this phase, the evaluation guidelines can be updated based on insights gained. Reviewers are re-trained according to the new guidelines, and the evaluation process is repeated until consensus is reached, such as achieving a Cohen's kappa value of 0.7 or above.

**Scoring and Review Phase**

Once consensus is achieved, the final score for each dimension is calculated using either mean or median scores, aggregating the evaluators' ratings. This scoring approach provides a comprehensive overview of the LLM's performance. To ensure a holistic evaluation, these human assessment results are compared with automatic evaluation metrics, such as the F1 measure and AUROC. This comparison benchmarks the human evaluation against machine-generated outputs, offering a rounded perspective on the strengths and limitations of the LLMs.

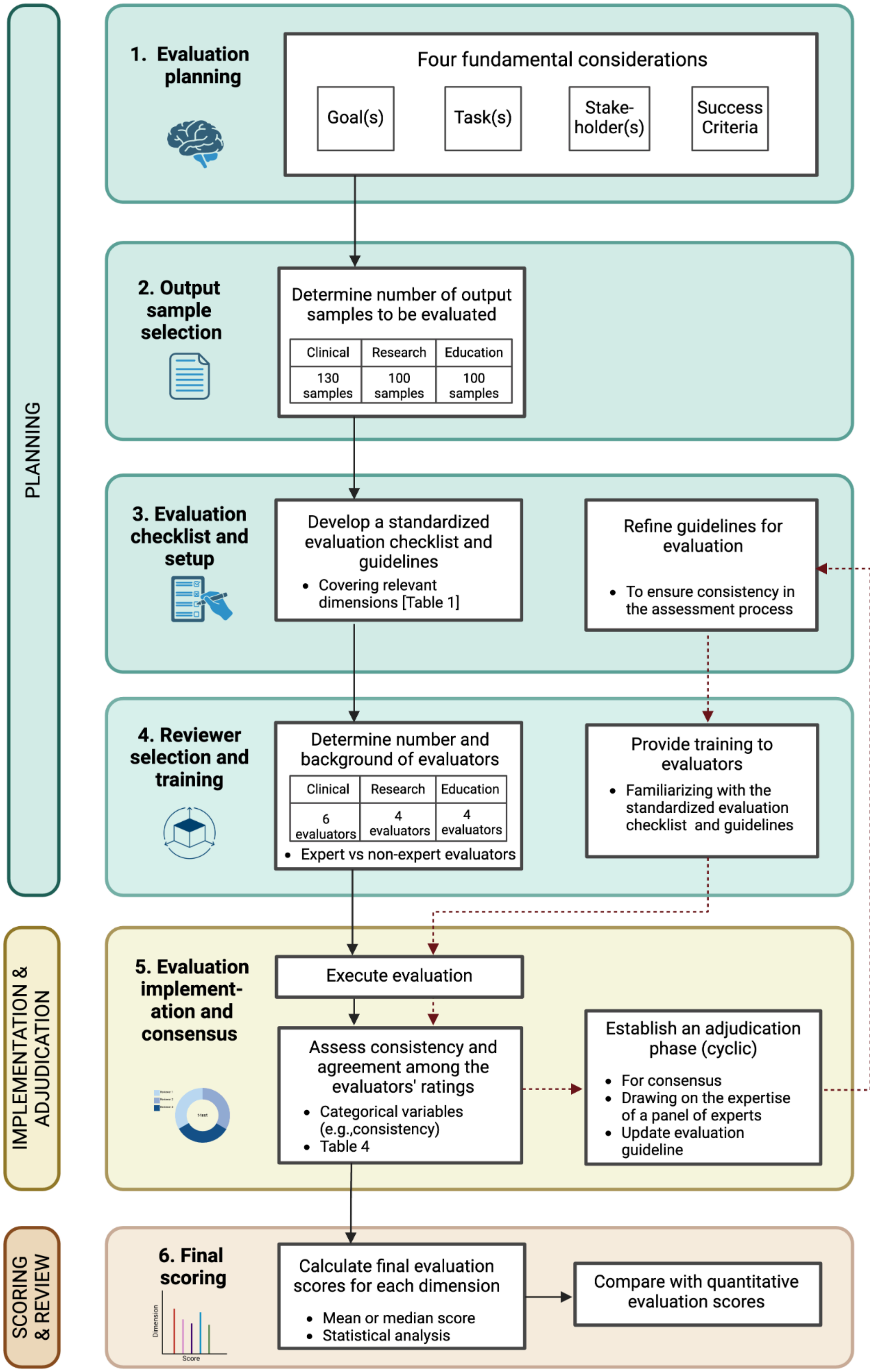

PLANNING
IMPLEMENTATION & ADJUDICATION
SCORING & REVIEW
1. Evaluation planning
Four fundamental considerations
Goal(s)
Task(s)
Stake-holder(s)
Success Criteria
2. Output sample selection
Determine number of output samples to be evaluated
Clinical
Research
Education
130 samples
100 samples
100 samples
3. Evaluation checklist and setup
Develop a standardized evaluation checklist and guidelines
• Covering relevant dimensions [Table 1]
Refine guidelines for evaluation
• To ensure consistency in the assessment process
4. Reviewer selection and training
Determine number and background of evaluators
Clinical
Research
Education
6 evaluators
4 evaluators
4 evaluators
• Expert vs non-expert evaluators
Provide training to evaluators
• Familiarizing with the standardized evaluation checklist and guidelines
5. Evaluation implement-ation and consensus
Execute evaluation
Assess consistency and agreement among the evaluators' ratings
• Categorical variables (e.g.,consistency)
• Table 4
Establish an adjudication phase (cyclic)
• For consensus
• Drawing on the expertise of a panel of experts
• Update evaluation guideline
6. Final scoring
Dimension
Score
Calculate final evaluation scores for each dimension
• Mean or median score
• Statistical analysis
Compare with quantitative evaluation scores

**Figure 8: The proposed human evaluation framework, delineating the multi-stage process for evaluating healthcare-related LLMs**

# 4. Case Studies

In this section, we provide two case studies of application of the QUEST framework in the Emergency Medicine speciality in a healthcare system to showcase the considerations needed for effective evaluation of LLMs in clinical use. Emergency departments (EDs) are the first point of contact for patients requiring urgent medical attention, requiring summarization of key patient information, generation of possible diagnoses and providing initial stabilization of patients with a wide variety of medical problems. The healthcare team needs to have a high level of expertise and make rapid management decisions. LLMs, by virtue of their ability to understand natural language, can be very useful to the ED team by improving the efficiency of triage workflow, making it quicker, accurate and free of fatigue, human error and biases. Figure 9 provides a side-by-side visual summary of how the QUEST framework is applied in two ED use cases: clinical note summarization and triage decision support. Details are provided below.

## 4.1: Human Evaluation of LLMs for Clinical Note Summarization in Emergency Department

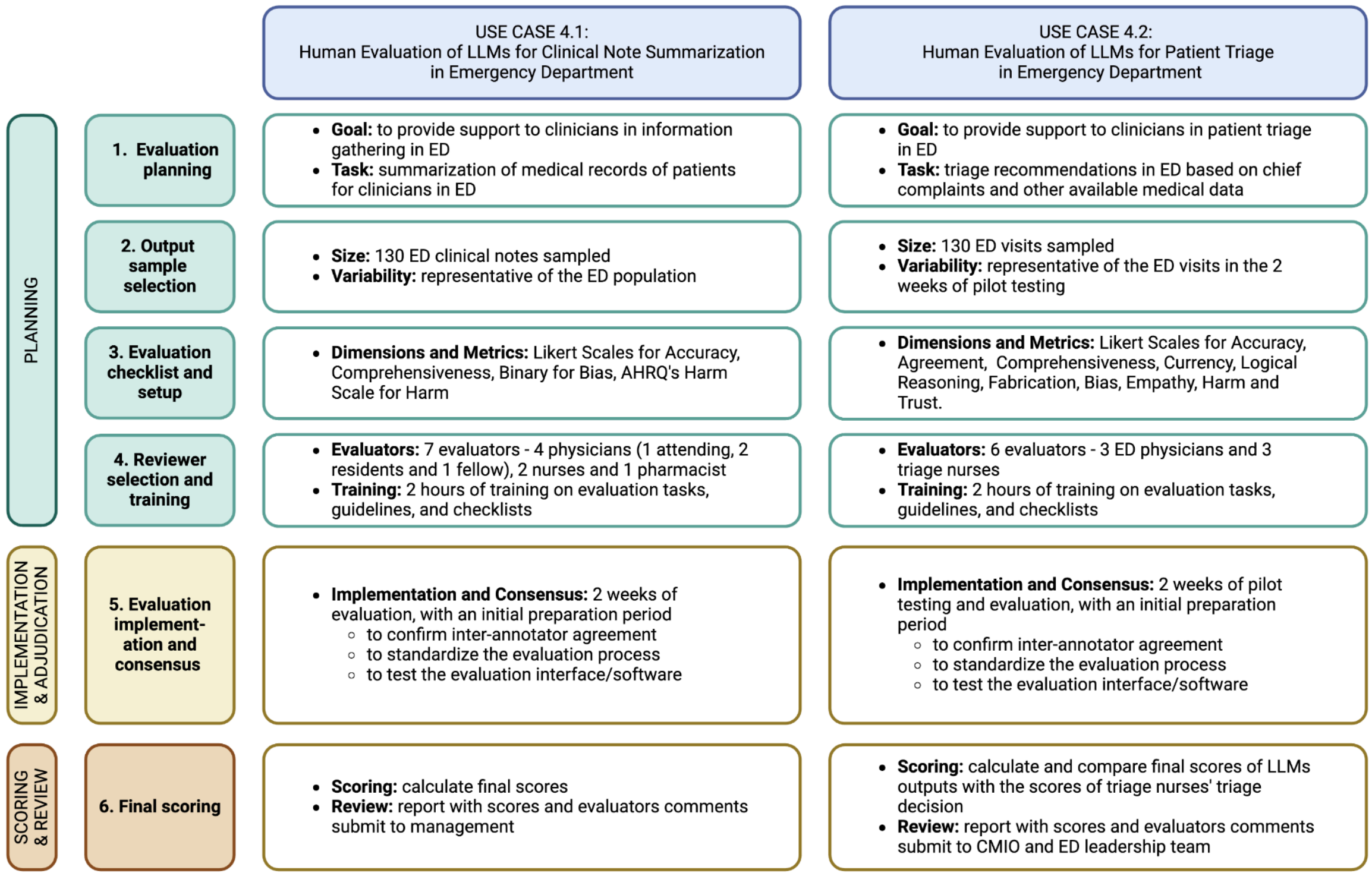

Figure 9: A visual summary of the application of QUEST framework in the cases of ED clinical note summarization and ED triage decision support in a healthcare system

Suppose a healthcare system wishes to evaluate a few LLMs for clinical note summarization to streamline clinical documentation and reduce physician burden. We will apply the QUEST framework to the human evaluation of LLMs in this application. First, during the Planning Phase, the team needs to consider four key elements:

1. Goals of the Model: The primary objective of implementing LLMs for clinical note summarization is to streamline the clinical documentation process, thereby significantly reducing the documentation burden on physicians.
2. Tasks Performed by the Model: The model will perform several critical tasks to achieve its objectives. It will extract key clinical information from detailed patient notes, including diagnoses, treatments, and outcomes. Following extraction, the model will generate concise and accurate summaries that retain all essential information. These summaries will be consistently formatted according to the healthcare system's documentation standards.
3. Stakeholders Involved: The primary users of the model are physicians who will benefit from reduced documentation time and improved workflow efficiency. Secondary users include nurses and allied health professionals who will refer to the summarized notes for patient care coordination. Patients are indirectly affected as improved documentation can lead to better quality of care and more efficient clinical workflows.

4. Criteria for Success: Success will be measured through several key criteria. Accuracy is paramount; the summaries must accurately reflect the key points of the original clinical notes without omitting critical information.

After defining these four considerations, the team will determine the sample size. Following the QUEST framework, they decided on a minimum sample size of 130 notes. Since the dataset lacks demographic information, they extracted such data from the notes where feasible to ensure sufficient variability that matches the population.

An important aspect of the QUEST framework is the dimensions and evaluation checklist. The committee identified dimensions most relevant to the emergency department (ED) settings where the model will be tested. Clinicians led the discussion, prioritizing Accuracy and Comprehensiveness under the principle “Quality of Information” while considering Currency less important. For metrics, they decided on a binary (presence or absence) metric for the “Bias” dimension and the AHRQ’s harm scale for the “Harm” dimension to capture ample details for categorizing the LLM outputs.

Recruiting evaluators for a clinical use case, as suggested by the QUEST framework, requires a minimum of seven evaluators. The committee determined that evaluators should have a suitable background reflective of the users, namely physicians, nurses, and other specialists. Since this use case is clinician-facing only, patient evaluators are not needed but would be included in other settings. The committee reached out to the ED department and recruited seven evaluators, including four physicians (one attending, two residents, and one fellow), two nurses, and one pharmacist.

Based on the QUEST framework, the evaluation involves five steps: developing the evaluation guideline, training evaluators, initial evaluation with a subset, discussion and adjudication, and fine-tuning the evaluation guideline. The committee developed an initial evaluation guideline and provided two hours of training to the evaluators. A sample of ten outputs with the original medical records was provided for the first round of evaluation. After discussing inter-annotator agreement to finalize and standardize the evaluation process, the guidelines were updated. The final evaluation was conducted by the seven evaluators over a two-week period. Evaluators' comments and scoring results were collected using an application built by the informatics team, and statistical analysis was performed by the committee. The final results report was then submitted to hospital management for an in-depth review before hospital-wide implementation.

## 4.2: Human Evaluation of LLMs for Patient Triage in Emergency Department

Suppose the Chief Medical Informatics Officer (CMIO) of a large academic health system wishes to evaluate LLMs as a decision support tool for ED teams in the triage of patients across all busy EDs in the health system. Figure 9 provides an example of how one could apply the QUEST framework to the human evaluation of LLMs in this application. During the Planning Phase, the team needs to consider four key elements:

1. Goals of the Model: The primary objective of implementing LLMs as a decision support tool is to enhance the efficiency and accuracy of patient triage in busy EDs.
2. Tasks Performed by the Model: The LLMs will use patients' chief complaints, relevant medical history, vital signs, and physical examination findings to make triage decisions (emergent, non-emergent, self-care at home) and provide recommendations for the next steps in management.
3. Stakeholders Involved: The primary users of the model are ED physicians, nurses, and triage staff, who will benefit from enhanced decision support and streamlined triage processes. Secondary stakeholders include patients, who will receive more timely and accurate triage, and hospital administration, which will benefit from improved ED efficiency and patient throughput.
4. Criteria for Success: Success will be measured through several key criteria, with accuracy in triage decisions being paramount. The model must provide reliable and timely triage recommendations that align with clinical assessments.

A total of 130 triage cases will be sampled and verified to be representative of ED visits for expert evaluation. Triage experts will perform their assessments and make triage decisions independently of the LLM output, serving as the gold standard. 3 experienced ED physicians and 3 nurses, different from the triage experts, will independently evaluate and compare the LLM outputs in terms of triage decisions with the gold standard and their own assessments. The core dimensions to be evaluated will include accuracy, agreement, comprehensiveness, currency, logical reasoning, fabrication, empathy, bias, harm, and trust, as outlined by the QUEST framework. Additionally, the evaluators will assess the LLMs' recommendations for the next management steps on a 5-point Likert scale (strongly agree, agree, neither agree nor disagree, disagree, strongly disagree).

The evaluations by the ED physicians will be completed using a standardized questionnaire (listed in Appendix Table A3). If there is any disagreement among the evaluators' results, adjudication will be performed until a consensus is reached. The final results of the evaluation will be collected and analyzed across various dimensions using appropriate statistical tests, and final evaluation scores for each dimension will be generated. The final results of the evaluation will be presented to the CMIO and ED leadership. This comprehensive evaluation will guide the decision on whether to implement the LLMs system-wide, ensuring it meets the healthcare system’s standards for accuracy, efficiency, and usability.

# 5. Discussion

LLMs have become integral to various clinical applications due to their ability to generate text in response to user queries. Despite their widespread use, the inner workings of these models remain opaque, in other words, they are still “black boxes”. The articles we reviewed reveal that evaluations of these “black box” models typically involve manual testing through human evaluation, which underscores a significant issue: the lack of traceability, reliability, and trust. Critical details such as the origins of the text sources, the reasoning processes within the generated text, and the reliability of the evidence for medical use are often not transparent. Furthermore, the traditional NLP evaluations, commonly used in well-defined tasks like Information Extraction (IE) and Question Answering (QA), prove to be suboptimal for assessing LLMs. This inadequacy stems from the novelty of the text generated by LLMs, which traditional NLP evaluation methods struggle to handle effectively. :As the use of LLMs in medicine increases, the need for appropriate evaluation frameworks which align with human values becomes more pronounced.

To address these challenges, we have proposed guidelines for human evaluation of LLMs. However, these too have limitations, constrained by the scale of human evaluation, the size of the samples reviewed, and the measures used, all of which can affect the depth and breadth of the assessments. Adding to these challenges is the predominance of proprietary models developed by major technology firms. The healthcare sector often faces constraints in computational resources, which limits the ability of informatics researchers to thoroughly study LLMs. This situation calls for a collaborative effort among the medical community, computer scientists, and major tech companies to develop comprehensive evaluation methods that can improve the quality and reliability of LLMs for clinical use. Our hope is to bridge these gaps and foster a synergy that could lead to more robust, transparent, and accountable LLMs in healthcare, ensuring they meet the high standards necessary for clinical application.

## Limitations

This literature review has a few limitations. First, this review may have missed relevant articles published after February 22, 2024. Since this field is rapidly evolving, there are likely significant advancements and new findings that have emerged since then. Second, the review is limited to articles written in English. Articles written in other languages may also provide valuable information and insights. Third, the search strings and databases we used for this review might not have been comprehensive enough, potentially introducing bias into the findings. Fourth, this review does not consider articles that utilize LLMs in other domains than healthcare, which may contain valuable insights regarding human evaluation.

The QUEST framework is developed based on the literature review and has several potential limitations as well. First, the implementation of QUEST framework may still vary across different clinical specialties and institutions. While we propose the framework to be generally applicable in various domains in healthcare and to best accommodate the varying limiting factors in different application settings, i.e. clinical, education, and research, constraints among subspecialties and/or institutions may pose challenges to their adherence to the framework and limit the actual outcome of standardization efforts advocated in this framework. Institutional policies, technology infrastructures, resources, and the availability of trained informatics personnel can vary widely, further complicating the implementation. Second, the dimensions of QUEST framework, although comprehensive, might not fully capture the nuances of every particular use case or scenario. The evaluation of LLMs is inherently task-specific in healthcare. Users should view this framework as a foundational starting point rather than a definitive solution. It is crucial to think critically about how this framework can be adapted and applied to meet the unique demands of their specific clinical applications. We encourage readers to consider applying a combination of our framework and specific frameworks as listed in section 3.4 depending on the use case. Third, the QUEST framework focuses only on human evaluation and does not consider automatic evaluation. While human evaluation provides critical insights into qualitative aspects of performance, automatic quantitative evaluation methods can offer valuable metrics and scalability. We fully acknowledge the utility that automatic evaluation can bring to any experiment or deployment of LLMs. Therefore, striking a balance between human evaluation and automatic quantitative evaluation is essential to uncover insights potentially missed by either approach.

## Future Research Directions

There are opportunities in unifying automatic and human evaluation across various applications in the healthcare domain. While there are efforts in aligning and improving the correlation between automatic evaluation metric and human evaluation such as Krishna et al. and Moramarco et al. [108,109], they are limited to specific applications and tasks such as clinical note summarization and note generation.

There are also technical advancements of multimodal LLMs and text-to-image foundational models such as CLIP [110]. These models can be useful in various healthcare applications, such as generating synthetic medical images (text-to-image), performing clinical diagnosis (image-to-text or recordings-to-text) and their potentials warrant a related but separate research.

Further, the ultimate goal of any implementation of LLMs is to improve actual patient outcomes or scientific understanding, and how to best connect the evaluation to reflect the impact of LLMs in achieving these goals is left as future work. While our evaluation framework aims to be futureproof, emerging techniques in LLMs and AI might require add-ons to our framework in the future. For example, emerging automatic and/or human evaluation methods with LLMs in the general domain may be helpful in the development of automatic and/or human evaluation healthcare domain, and vice versa.

Last, how to best apply LLMs and other emerging technology to automate or make human evaluation more scalable and low resources remains a research question. There are already attempts in applying LLMs to complement human evaluation but to a limited success.

# Acknowledgements

Research reported in this article was supported by the University of Pittsburgh Momentum Funds, and the National Institutes of Health awards UL1 TR001857, U24 TR004111, and R01 LM014306. The sponsors had no role in study design; in the collection, analysis, and interpretation of data; in the writing of the report; and in the decision to submit the paper for publication.

# Contributions

T.Y.C.T. and S.S. conceptualized, designed, and organized this study, analyzed the results, and wrote, reviewed, and revised the paper. S.K., A.V.S., K.P., K.R.M., H.O., and X.W. analyzed the results, and wrote, reviewed, and revised the paper. S.V., S.F., P.M., G.C., C.S., and Y.P. wrote, reviewed, and revised the paper. Y.W. conceptualized, designed, and directed this study, wrote, reviewed, and revised the paper.

# Declaration of conflicting interests

P.M. has ownership and equity in BrainX, LLC, Y.W. has ownership and equity in BonafideNLP, LLC, and S.V. has ownership and equity in Kvatchii, Ltd., READE.ai, Inc., and ThetaRho, Inc . The other author(s) declared no potential conflicts of interest with respect to the research, authorship, and publication of this article.

# Appendix

Figure A1 - Distribution of LLMs

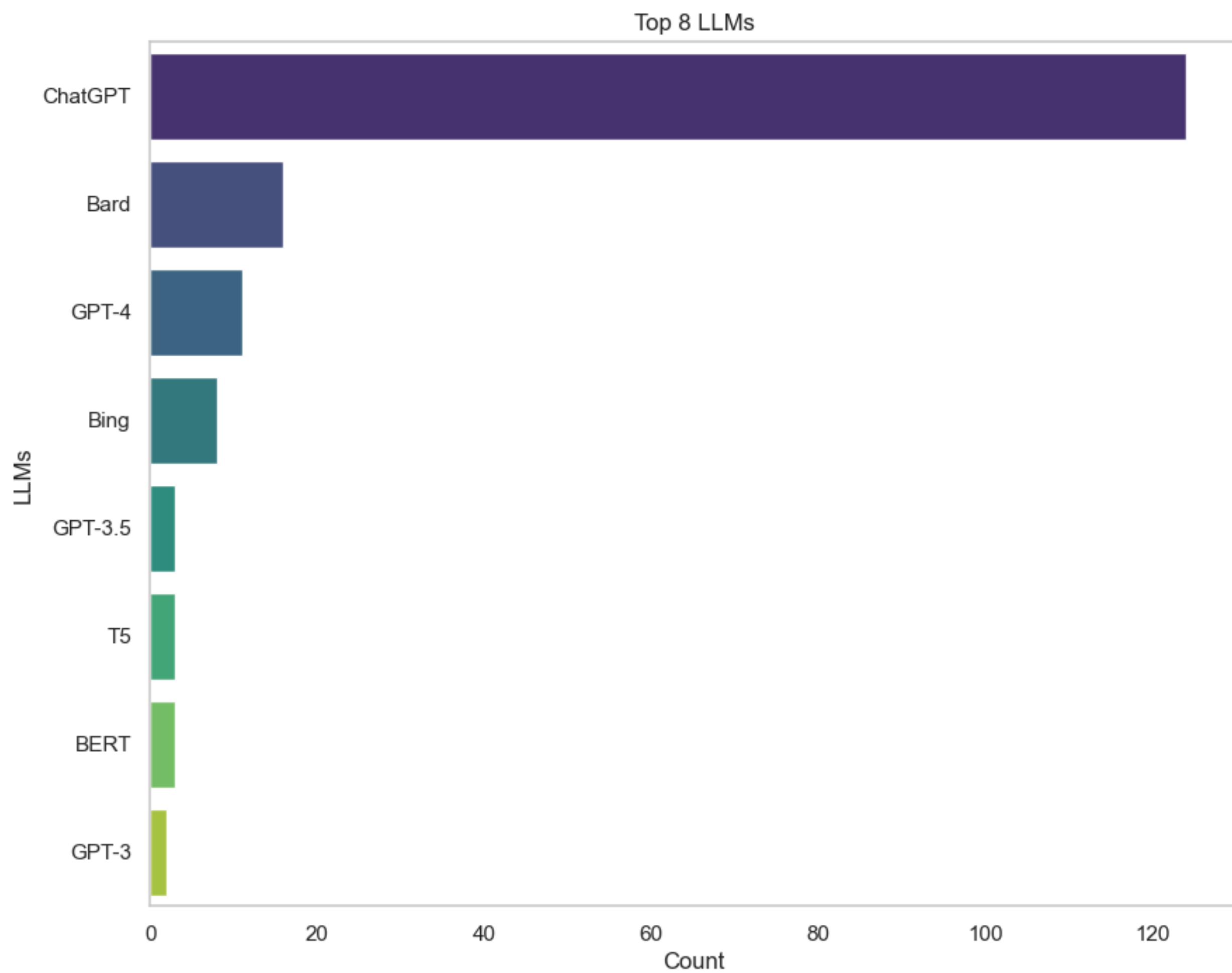

Majority of studies reviewed applied GPT family models developed by OpenAI, reflecting the popularity among the general public. It is noteworthy that open source models , such as Llama by Meta are not among the top of the list.

Figure A2 - Comparison of LLMs

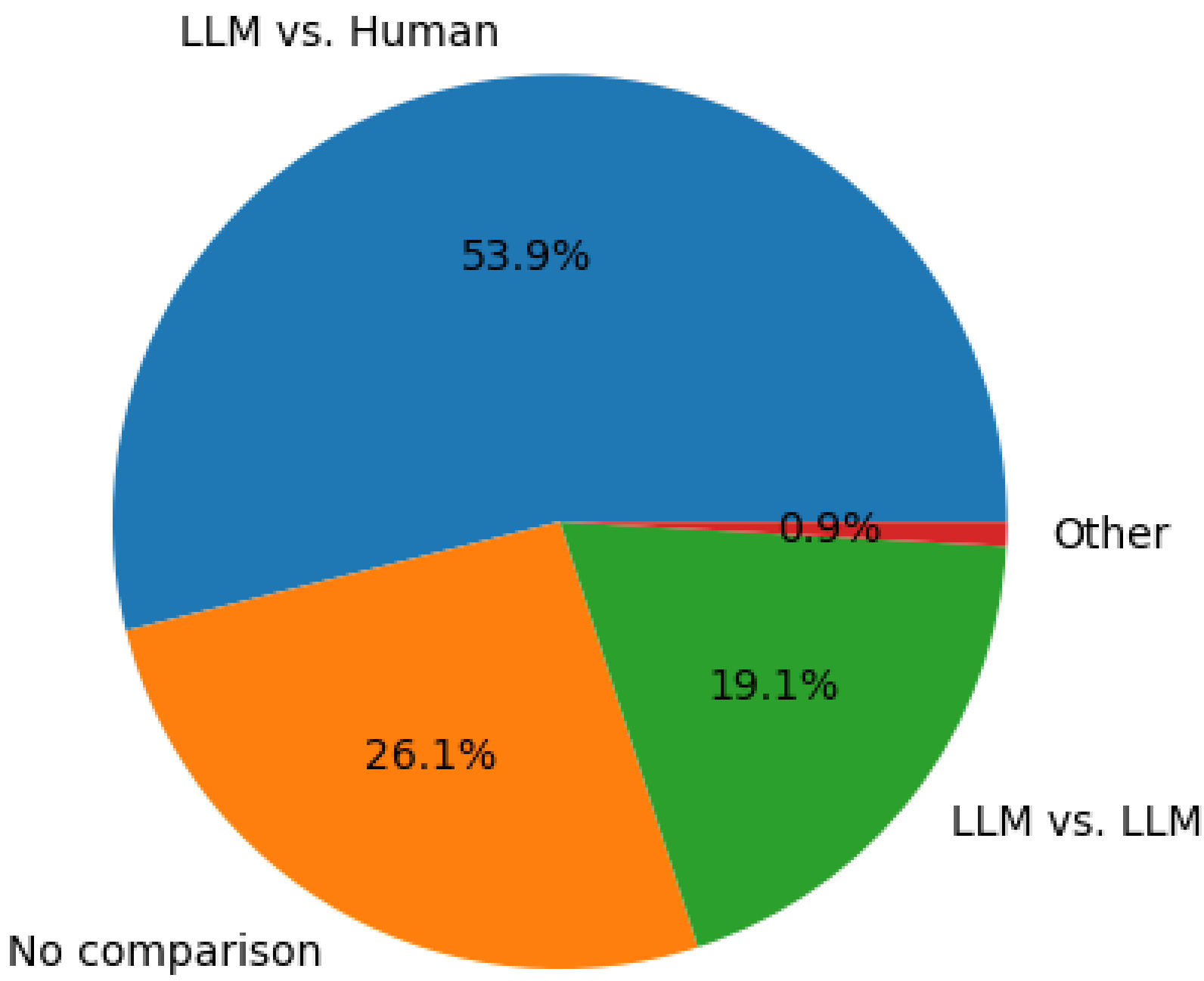

Table A3 - Evaluation Questions for Case Study: Human Evaluation of LLMs for Patient Triage in Emergency Department

| |
|---|
| What is the Medical Record Number (MRN) of the case? |
| What was the triage decision of the ED Triage Nurse? |
| Do you agree with ED Triage Nurse's triage decision? |
| What was the triage decision of the LLM from its output? |
| Do you agree with the LLM's triage decision? |
| Was the LLM's output regarding patient's management (diagnostic plan and treatment) accurate (factually correct, precise and free of errors)? |
| Was the LLM's output regarding patient's management (diagnostic plan and treatment) in agreement with established facts and guidelines? |
| Was the LLM's output regarding patient's management (diagnostic plan and treatment) comprehensive (providing a complete overview and detailed insights)? |
| Was the LLM's output regarding patient's management (diagnostic plan and treatment) consistent with most recent literature and updated guidelines about the disease? |
| Did LLM's output regarding patient's management (diagnostic plan and treatment) apply logical reasoning to generate the response? |
| Was the LLM's output regarding patient's management (diagnostic plan and treatment) fabricated (contained made up information, non-existent facts, omitting or changing critical facts)? |
| Was the LLM's output regarding patient's management (diagnostic plan and treatment) empathetic (recognized emotions and tone in the input and reflected consideration in the output)? |
| Was the LLM's output regarding patient's management (diagnostic plan and treatment) free of bias (racial and gender)? |
| Was the LLM's output regarding patient's management (diagnostic plan and treatment) harmful (spreading misinformation, adversely affecting users)? |
| Was the LLM's output regarding patient's management (diagnostic plan and treatment) trustworthy (confidence in the accuracy, safety and fairness of the LLM output)? |